%% file: main.tex
\documentclass[runningheads]{llncs}

\usepackage{eccv}

\usepackage{eccvabbrv}

\usepackage{graphicx}
\usepackage{booktabs}

\usepackage[accsupp]{axessibility}  

\usepackage{tcolorbox}
\usepackage{pifont}
\usepackage{longtable}
\usepackage{makecell}
\usepackage{multirow} 
\usepackage{array}
\usepackage{listings}
\newcommand{\NICKNAME}{\textsc{MemLearner}} 
\newcommand{\nickname}{\NICKNAME} 

\usepackage{hyperref}

\usepackage{orcidlink}

\begin{document}

\title{\nickname: Learning to Query Context Memory for Video World Models} 

\titlerunning{\nickname: Learning to Query Context Memory}

\author{Jiwen Yu\inst{1}\textsuperscript{\ensuremath{\ddagger}} \and
Jianxiong Gao\inst{2}\textsuperscript{\ensuremath{\ddagger}} \and
Jianhong Bai\inst{3}\textsuperscript{\ensuremath{\ddagger}} \and
Yiran Qin\inst{1} \and
Kaiyi Huang\inst{1}\textsuperscript{\ensuremath{\ddagger}} \and
Quande Liu\inst{4} \and
Xintao Wang\inst{4}\textsuperscript{\ensuremath{\dagger}} \and
Pengfei Wan\inst{4} \and
Kun Gai\inst{4} \and
Xihui Liu\inst{1}\textsuperscript{\ensuremath{\dagger}}}

\authorrunning{J.~Yu et al.}

\institute{The University of Hong Kong \and
Fudan University \and
Zhejiang University \and
Kuaishou Technology}

\maketitle
\begingroup
\renewcommand{\thefootnote}{}
\footnotetext{\textsuperscript{\ensuremath{\ddagger}}Work done during an internship at Kling Team, Kuaishou Technology. \textsuperscript{\ensuremath{\dagger}}Corresponding authors. Project page: \url{https://yujiwen.github.io/memlearner/}.}
\endgroup

\input{fig_tab_tex/teaser}

\input{sec/0_abstract}
\input{sec/1_intro}
\input{sec/2_related}

\input{sec/3_method}
\input{sec/4_exp}
\input{sec/5_conclusion}

\section*{Acknowledgements}
This work is supported by the Research Grant Council of Hong Kong through Early Career Scheme under grant 27214925. We thank Yao Teng, Xiaoshi Wu, and Haoran He for constructive discussions and suggestions.

%
%
\bibliographystyle{splncs04}
\bibliography{main}

\clearpage
\appendix
\renewcommand{\theHsection}{appendix.\arabic{section}}
\renewcommand{\theHsubsection}{appendix.\arabic{section}.\arabic{subsection}}
\input{tex/dataset}

\input{tex/model}
\input{tex/exp}

\input{tex/tab_fig}
\end{document}

%% file: fig_tab_tex/teaser.tex


\begin{figure}[h]
  \centering
  \vspace{-0.6cm}
  \includegraphics[width=1\linewidth]{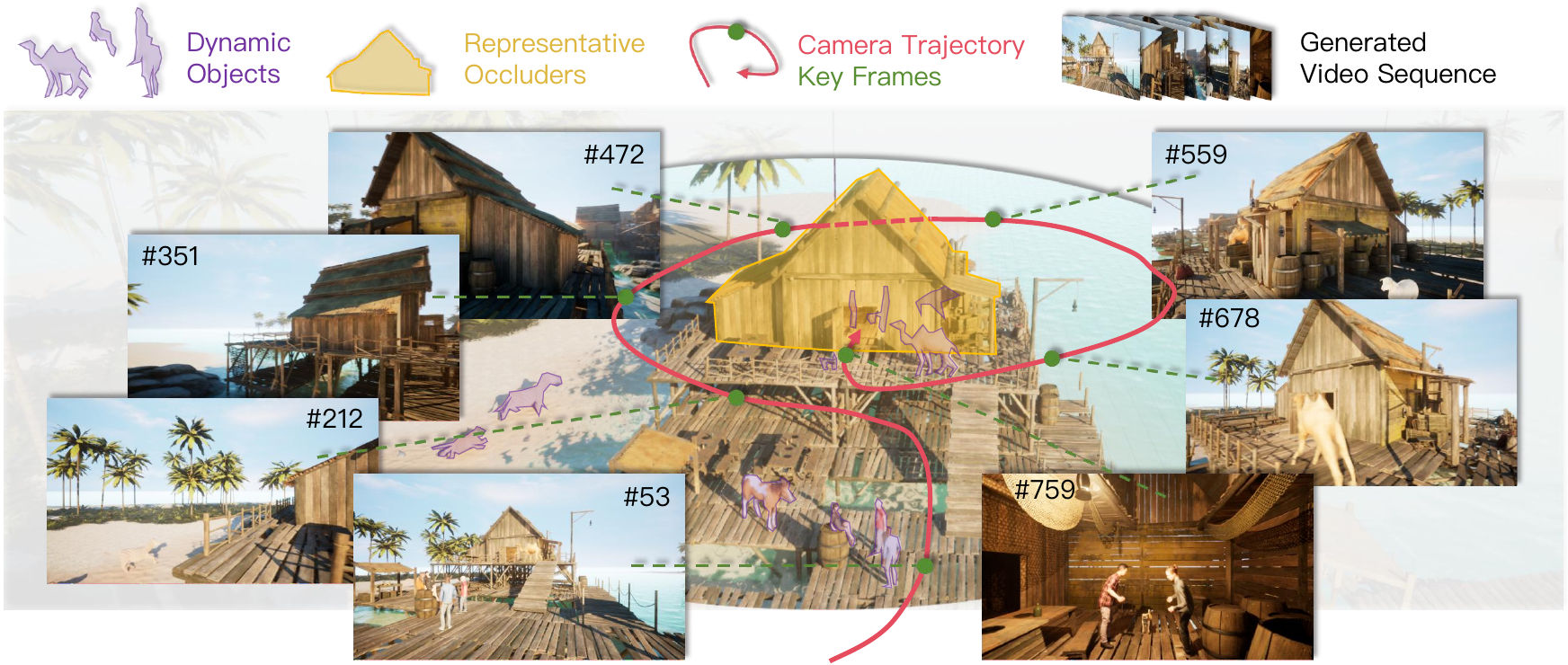}
  \caption{\textbf{Teaser Demonstration.} We propose a novel memory mechanism for video world models through learning-based adaptive context querying. Compared to prior rule-based context retrieval methods~\cite{cam,worldmem,vmem}, our approach handles scenes with occlusions and dynamic objects. This figure highlights dynamic objects, representative occluders, video generation trajectory, and key frames.}
  \vspace{-0.8cm}
\label{fig:teaser} 
\end{figure}


%% file: sec/0_abstract.tex
\begin{abstract}

Video World Models are interactive video generation models that predict future world states based on user actions and history video frames. A critical challenge in video world models is the lack of memory, causing inconsistent generated scenes over extended durations. Previous methods explored rule-based context frame retrieval as memory, but they fail to generalize in scenarios with scene occlusions and dynamic objects. We propose \textbf{\textsc{MemLearner}}, a learning-based adaptive context query method using query tokens to bridge context and predicted tokens. By leveraging the video generation model itself for context querying, \textsc{MemLearner} exploits pre-trained visual priors without training additional modules from scratch, and incorporates efficient strategies for training and inference. We collect a dataset of long videos with scene occlusions and dynamic objects, paired with camera pose annotations, and propose a multi-dataset training strategy leveraging both annotated rendered and unannotated real-world videos. Extensive experiments demonstrate that \textsc{MemLearner} significantly outperforms prior video world models in terms of scene consistency and memory, particularly under challenging occlusion and dynamic scenarios.
\keywords{Video World Model \and Interactive Video Generation \and Video Diffusion Model}
\end{abstract}

%% file: sec/1_intro.tex
\section{Introduction}
\label{sec:intro}

World Models~\cite{worldmodel} understand and simulate world dynamics by taking historical world states and interactive actions as input to predict new states. World states can be represented as language~\cite{gpt3}, latent representations~\cite{worldmodel,vjepa2,dino-wm}, 3D/4D~\cite{wonderworld,Worldlabs}, or videos~\cite{sora,genie3}. Among these, Video World Models are particularly promising~\cite{video-new-language,igv-position,igv-survey} as videos photorealistically capture real-world dynamics, and abundant internet video data offers significant scaling potential. Despite significant progress made in generating short video clips, Video World Models still face significant challenges of scene consistency over extended durations, caused by insufficient memory mechanisms.

The memory issue in Video World Models arises when later generated scenes become inconsistent with earlier ones due to limited context windows.  
Existing methods address this challenge by different memory representations: 3D reconstruction~\cite{wonderjourney,wonderplay,viewcrafter,see3d,gen3c,spmem}, compressed feature representations~\cite{3d-persistent-wm, gamegan,framepack,ssvwm}, and retrieving key context frames~\cite{cam, worldmem, vmem, vrag}.
Context retrieval is particularly promising as it eliminates additional costs and potential errors caused by 3D reconstruction or history compression.

However, existing context retrieval methods are rule-based, relying on FOV overlap~\cite{cam,worldmem} or point cloud estimation and surfel matching~\cite{vmem}, facing several fundamental limitations in complex scenes with occlusions and dynamic objects. 
For example, FOV-based context retrieval~\cite{cam} cannot account for occluding walls between camera views, point-cloud-based retrieval methods cannot accurately reconstruct moving objects. Moreover, these hard-coded retrieval rules fail to take generalized and dynamic environments with state changes into consideration.
\textbf{These limitations motivate a paradigm shift: rather than using hand-crafted rules to retrieve context, we propose a new memory mechanism that enables the network to learn to adaptively query information from historical frames through end-to-end training.}

\input{fig_tab_tex/arch_clarify}
We formulate history-conditioned video generation as predicting future frames based on historical context frames. To design a learning-based context query method, we introduce query tokens ($\mathbf{Q}$ tokens) as an information bridge between context tokens ($\mathbf{C}$ tokens) and predicted tokens ($\mathbf{P}$ tokens) (Fig.~\ref{fig:arch_clarify} (a)): $\mathbf{Q}$ tokens attend to $\mathbf{C}$ tokens to adaptively extract context information, while $\mathbf{P}$ tokens attend to $\mathbf{Q}$ tokens as the generation condition.
A key design choice is to leverage the video generation model itself for context querying, rather than introducing a separate module. Specifically, we feed all $\mathbf{C}$, $\mathbf{Q}$, and $\mathbf{P}$ tokens together into the video generation model (Fig.~\ref{fig:arch_clarify} (c), detailed in Sec.~\ref{subsec:learn-to-query}), exploiting the model's pre-trained visual priors without additional scratch-trained modules. To further reduce the computational cost of context querying over long video sequences, we propose two efficient strategies for training and inference (Sec.~\ref{subsec:efficient}).

Training learning-based context querying requires a long video dataset with occlusion, dynamic objects, camera pose annotations, and sufficient diversity. However, no existing datasets~\cite{cam,sekai,spatialvid,omniworld} fully meet these criteria. Real-world videos (e.g., YouTube) provide diverse and dynamic content but lack precise camera pose annotations, while rendered videos (e.g., Unreal Engine) offer accurate poses but suffer from limited visual realism and diversity.
To address this, we make two contributions. First, we collect a rendered dataset with customized occlusions and dynamic objects (Sec.~\ref{subsec:dataset_collection}). 
Second, we propose a multi-dataset training strategy that assigns a dedicated camera encoder to each dataset type, where unannotated data is fed with zero camera parameters. By isolating distinct annotation qualities into separate encoders, the model simultaneously leverages the strengths of diverse data sources without mutual interference (Sec.~\ref{subsec:training_strategy}).

Our contributions can be summarized as follows:
\begin{itemize}
    \item We propose \textsc{MemLearner}, a learning-based adaptive context query method that utilizes query tokens to extract valid memory from context tokens, enabling memory-augmented Video World Models with improved consistency and generalization.
    \item We collect a customized video world model dataset based on Unreal Engine, incorporating occlusion and dynamic objects, and propose a multi-dataset training strategy to leverage the advantages of both rendered and real videos.
    \item Extensive experiments demonstrate that \textsc{MemLearner} significantly outperforms previous Video World Models in terms of scene consistency and memory, particularly under occlusion and dynamic object scenarios.
\end{itemize}

%% file: fig_tab_tex/arch_clarify.tex
 \begin{figure}[!tbp]
  \centering
  \includegraphics[width=1\linewidth]{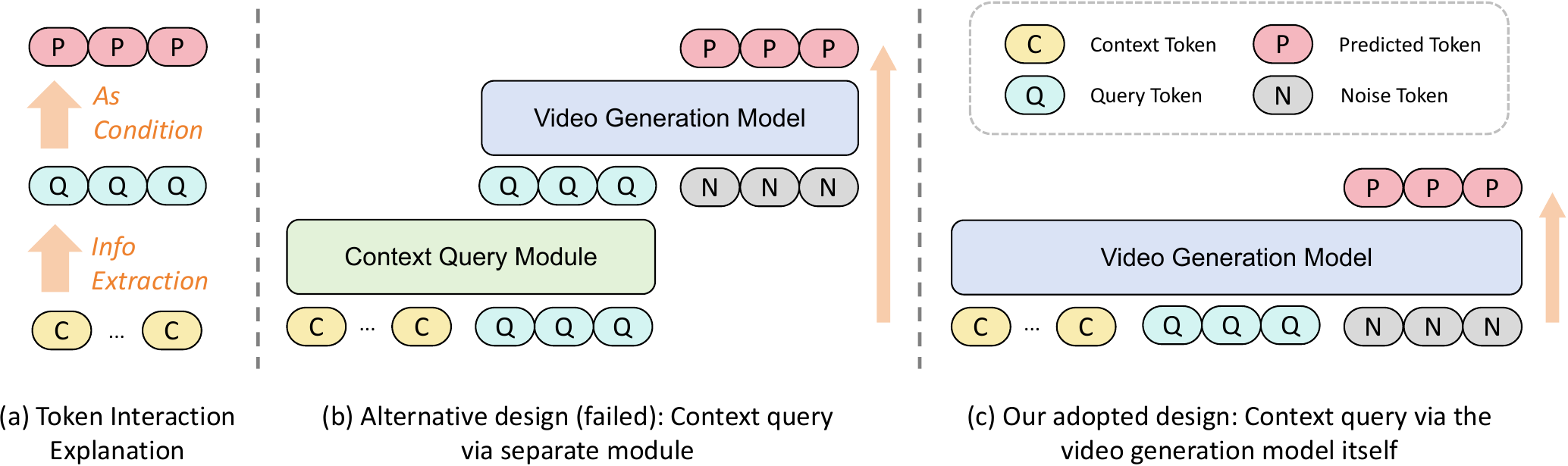}
  \caption{\textbf{Architecture clarification.} (a) Interaction mechanism among $\mathbf{C}$, $\mathbf{Q}$, and $\mathbf{P}$ tokens; (b)\&(c) Two designs for context querying, where the alternative design in (b) fails in experiments (Sec.~\ref{subsec:exp_comp}), and the adopted design in (c) leverages the prior knowledge of the video generation model itself and performs effectively.}
  \vspace{-0.2cm}
\label{fig:arch_clarify} 
\end{figure}

%% file: sec/2_related.tex
\section{Related Work}
\label{sec:related}

\subsection{Video World Models}

\paragraph{World Model.} 
The World Model~\cite{worldmodel} refers to a system that takes historical world state and current action as input to predict future world state. Its primary purpose is to understand and simulate the evolving process of the world. The world state can be represented in various forms, such as language~\cite{gpt3}, semantic representations~\cite{dino-wm,vjepa2}, 3D/4D~\cite{wonderworld,Worldlabs}, and videos~\cite{sora,genie3}. Recently, due to video photorealism, large data volumes, and breakthroughs in video generation models, Video World Models are considered a promising path to World Models~\cite{video-new-language,igv-position,igv-survey}.

\paragraph{Video Generation Models.}
Mainstream video generation models now use the Diffusion Transformer (DiT)~\cite{dit,sora} architecture, enabling highly realistic video creation~\cite{sora,cogvideox,kling,veo2,runway,vidu,wan,hunyuanvideo}. Other architectures include next-token prediction~\cite{videogpt,videopoet,emu3,emu35} and hybrid approaches~\cite{cdf,nova,mar}, but fall short of DiT in generation quality. Interactive control in video generation includes camera pose~\cite{cameractrl,motionctrl,recammaster}, trajectory~\cite{motionctrl,3dtrajmaster}, and action control~\cite{gamefactory,irasim,gamengine}, with applications in film production, game video control, and robotic simulation. Streaming video generation conditions on previously generated frames to create new content, with both frame-by-frame~\cite{cdf,dfot,framepack} and chunk-by-chunk methods~\cite{cam,gamefactory,magi1} primarily using DiT. Long video generation suffers from error accumulation, which several methods address~\cite{self-forcing,self-forcing++,longlive,rolling-forcing,svi}. Since our work focuses on learnable memory rather than ultra-long generation, we do not discuss error accumulation.

\subsection{Memory Mechanism for Video World Models}
Memory capability in video world models refers to generating consistent long videos, especially when revisiting scenes or objects. It is a fundamental prerequisite for planning, interaction and state modeling, as these core capabilities rely on retaining past visual memories.
However, many methods struggle with this~\cite{gamengine,oasis,dfot,wham}, due to limited context windows that cannot retain sufficient historical information.
To address this, additional conditional information is necessary. Existing approaches fall into three paradigms: (1) 3D as memory~\cite{wonderjourney,wonderplay,viewcrafter,see3d,gen3c,spmem}: reconstructing 3D representations from historical frames and rendering new initial frames as conditions; (2) Feature as memory~\cite{3d-persistent-wm,gamegan}: extracting semantic features from historical frames or maintaining learnable features injected into the generation model; (3) Context as memory~\cite{cam,worldmem,vmem,ssvwm,vrag,framepack,relic,worldplay}: directly using historical frames as conditions. Among these, context memory is the most straightforward. However, incorporating all historical contexts introduces prohibitive computational overhead.


Some works propose context retrieval to select historical contexts relevant to current generation. Conditioning on only retrieved frames significantly reduces computational overhead. Existing methods typically rely on \textbf{rule-based retrieval}~\cite{cam,worldmem,vmem}, which lack generalization across videos and scenes. We propose a \textbf{learning-based context query method} enabling the model to learn to query useful historical information, thereby achieving learnable memory.

\subsection{Learnable Query Tokens in Various Domains}

Learnable query tokens have been widely adopted for compressing visual context across various domains:
In multimodal language models, Perceiver Resampler~\cite{flamingo} and Q-Former~\cite{blip2} use learnable queries to aggregate visual features for language model consumption;
In video understanding, adaptive frame selection methods~\cite{adaframe,buch2025flexible,arnab2025temporal} employ learnable strategies to identify informative frames for recognition and reasoning tasks.
Our work addresses a fundamentally different problem: memory issue in video world models, where query tokens serve as an information bridge between context tokens and predicted tokens, adaptively extracting fine-grained, token-level information across multiple historical frames to support consistent video generation under occlusion and dynamic scenarios.

%% file: sec/3_method.tex
\section{\textsc{MemLearner}}
\label{sec:method}

\subsection{Preliminaries}
\label{subsec:preliminaries}
\input{fig_tab_tex/dit}

Our approach is built upon a latent video diffusion model with a causal 3D VAE~\cite{vae} and Diffusion Transformer (DiT)~\cite{dit}. As shown in Fig.~\ref{fig:dit}, each DiT block sequentially contains spatial (2D) attention, spatial-temporal (3D) attention, cross-attention, and feed-forward networks (FFN). The 3D VAE encoder compresses a video frame sequence $\mathbf{x}$ temporally and spatially into latent representations $\mathbf{z} = E(\mathbf{x})$, and the DiT is trained to predict noise scaled and added to $\mathbf{z}$ via a standard diffusion objective.
We integrate camera control~\cite{motionctrl,recammaster} by injecting camera poses $\mathbf{cam} = [R, t] \in \mathbb{R}^{f \times (3 \times 4)}$ into the model via a one-layer MLP encoder $\mathcal{E}_c(\cdot)$, added to features between the 2D and 3D attention layers. Note that camera control only guides the trajectory of generated videos; our learning-based context query method does not rely on camera poses, as validated in Sec.~\ref{subsec:ablation}.
For long video generation, we follow a chunk-by-chunk autoregressive paradigm: historical context frame latents $\mathbf{h}$ are concatenated with the to-be-generated latents $\mathbf{z}$ along the frame dimension and fed jointly into the model, with the diffusion loss applied only to $\mathbf{z}$. This preserves the model's generative priors without architectural modifications.

\subsection{Learning to Query Context}
\label{subsec:learn-to-query}
\paragraph{Introducing Adaptive Query Tokens.}
Our key insight is that different predicted frames require different guidance information from context frames, and even within the generation of a single frame, different denoising stages of the diffusion process need to emphasize different aspects of historical information. We verify this via attention similarity analysis in Appendix~\ref{sec:attn_vis}: query tokens exhibit markedly different attention distributions across predicted frames and diffusion timesteps, with early timesteps attending more broadly to context while later timesteps focus on fine-grained local correspondences.
Naive history compression or rule-based keyframe retrieval fail to tackle complex scenarios such as occlusion and dynamic objects. To enable adaptive memory, we introduce learnable query tokens $\mathbf{Q}$ that bridge context and generation. These query tokens dynamically extract relevant information from context tokens $\mathbf{C}$ and guide the generation of predicted tokens $\mathbf{P}$. Importantly, this querying mechanism can be learned end-to-end via indirect supervision from the diffusion loss on $\mathbf{P}$ alone, without explicit supervision for a dedicated querying module.

\paragraph{Architecture Design Insights.}
The intuitive design introduces an additional context query module, as in Fig.~\ref{fig:arch_clarify} (b). However, our experiments show that when jointly training the architecture, the from-scratch context query module fails to learn useful information. Attention similarity analysis (Appendix~\ref{sec:attn_vis}) confirms that this module produces near-zero similarity between query and context tokens, indicating a failure to establish meaningful context modeling; this in turn hinders gradient propagation to the video generation model, which consequently ignores the module's output and degrades to a text-to-video model (Sec.~\ref{subsec:exp_comp}).
As in Fig.~\ref{fig:arch_clarify} (c), our design avoids a separate from-scratch module and instead leverages the pre-trained video generation model itself for context querying. This offers clear advantages: (1) it exploits the model's prior knowledge, reducing data and compute requirements; (2) end-to-end training naturally learns context query capability without requiring additional input-output design or supervision losses for a separate module.

\paragraph{Architecture Details.}
Our work is based on a latent video diffusion model. We denote the video latent tokens as $Z = \{ \mathbf{C}, \mathbf{Q}, \mathbf{P} \}$, where $\mathbf{C}$, $\mathbf{Q}$ and $\mathbf{P}$ denote context, query, and predicted tokens, respectively. The diffusion model input is the noisy token at timestep~$t$, written as $Z_t = \{ \mathbf{C}, \mathbf{Q}, \mathbf{P}_t \}$, where the context tokens and query tokens remain unperturbed, while the predicted tokens are noised only at the input (and not at any subsequent layer) via randomly sampled Gaussian noise to obtain $\mathbf{P}_t$, with the scaled and added noise being $\epsilon^P \sim \mathcal{N}(\mathbf{0}, \mathbf{I})$.
The training loss for the entire architecture is:
\begin{equation}
    \mathcal{L}(\theta) = \mathbb{E} [ || \epsilon_{\theta}( Z_t, \mathbf{cam}, \mathbf{p}, t) || - \epsilon^P ],
\end{equation}
where $\theta$ denotes all learnable parameters. Note that supervision is applied only to the noise predicted on predicted tokens.
We concatenate $\mathbf{C}$, $\mathbf{Q}$ and $\mathbf{P}$ along the frame dimension and input them into the video DiT, where different tokens interact with each other in the 3D attention.
For a 3D attention layer with input $F_{in}$ and output $F_{out}$, we have $F_{in} = \{ \mathbf{C}, \mathbf{Q}, \mathbf{P} \}$ and $F_{out} = \{ \mathbf{C}_{out}, \mathbf{Q}_{out}, \mathbf{P}_{out} \}$.
For brevity, we use $\mathbf{C}$, $\mathbf{Q}$, and $\mathbf{P}$ to denote the features corresponding to different tokens.
We define the linear computations in the attention block as $q(\cdot)$, $k(\cdot)$, $v(\cdot)$, and $o(\cdot)$,
which correspond to the operations for query, key, value, and output. A standard 3D attention computation is:
\begin{equation}
\begin{aligned}
    F_{out} &= F_{in} + o(\mathbf{sm}(q(F_{in}) k(F_{in})^\top ) v(F_{in}) ),\\
    &= F_{in} + g(F_{in}, F_{in}, F_{in}),
    \label{eq:all_compute}
\end{aligned}
\end{equation}
where $\mathbf{sm}(\cdot)$ denotes the softmax operation, and $g(\cdot, \cdot, \cdot)$ is a shorthand notation indicating the inputs for query, key, and value operations, respectively. Considering that the context frame length may be very large, this computation in Eq.~\ref{eq:all_compute} is expensive and inefficient.

\subsection{Efficient Strategies}
\label{subsec:efficient}
\input{fig_tab_tex/framework}
We present two simple yet effective strategies to improve the efficiency of Eq.~\ref{eq:all_compute} during both training and inference.

\paragraph{Strategy \#1: Query Only in Early Layers.} Information querying is analogous to encoding, which requires fewer parameters and computations than the generation model (e.g., video VAEs use far fewer parameters than video generation models). Therefore, context querying can be performed only in the early shallow layers, which our experimental results validate. Specifically, assuming the diffusion transformer has $n+m$ layers in total, we divide them into two types as shown in Fig.~\ref{fig:framework} (a): $n$ shallow Query Layers near the input and $m$ deep Generative Layers near the output, where $n\ll m$. Interactions among $\mathbf{C}$, $\mathbf{Q}$, and $\mathbf{P}$ occur only in Query Layers, while Generative Layers process only $\mathbf{Q}$ and $\mathbf{P}$.

\paragraph{Strategy \#2: Exclude Unnecessary Computations.} We exclude unnecessary attention computations by retaining only three essential patterns as shown in Fig.~\ref{fig:framework} (b): (1) $\mathbf{Q}$ as queries attending to $\mathbf{P}$ as keys/values, enabling $\mathbf{Q}$ to understand what information to extract from context based on the prediction target; (2) $\mathbf{Q}$ as queries attending to $\mathbf{C}$ as keys/values, enabling $\mathbf{Q}$ to extract information from context; (3) $\mathbf{P}$ as queries attending to $\mathbf{P}$ and $\mathbf{Q}$ as keys/values, enabling $\mathbf{P}$ to extract information from both. All other attention computations are excluded, particularly those where $\mathbf{C}$ serve as queries, significantly reducing computational overhead. We provide experimental analysis of performance under different attention patterns in Sec.~\ref{subsec:ablation}.

Combining the two strategies, we present the revised Eq.~\ref{eq:all_compute}. For Query Layers:
\begin{equation}
    \begin{aligned}
        \mathbf{C}_{out}&=\mathbf{C},\\
        \mathbf{Q}_{out}&=\mathbf{Q}+g(\mathbf{Q}, \{\mathbf{C}, \mathbf{P}\}, \{\mathbf{C}, \mathbf{P}\}),\\
        \mathbf{P}_{out}&=\mathbf{P}+g(\mathbf{P},\{\mathbf{P}, \mathbf{Q}\}, \{\mathbf{P}, \mathbf{Q}\}). \\
    \end{aligned}
\end{equation}
For Generative Layers, features of $\mathbf{C}$ are removed:
\begin{equation}
    \begin{aligned}
        \mathbf{Q}_{out}&=\mathbf{Q}+g(\mathbf{Q}, \mathbf{P}, \mathbf{P}),\\
        \mathbf{P}_{out}&=\mathbf{P}+g(\mathbf{P}, \{\mathbf{P}, \mathbf{Q}\}, \{\mathbf{P}, \mathbf{Q}\}). \\
    \end{aligned}
\end{equation}



\section{Dataset}
\label{sec:dataset}

\subsection{Dataset Collection}
\label{subsec:dataset_collection}
To validate our method, we require a dataset with specific characteristics: long videos, accurate per-frame camera annotations, occlusion relationships, dynamic objects, and sufficient diversity. However, existing long video datasets do not meet these requirements, as shown in Tab.~\ref{tab:dataset_comparison}. To address this, we collect a rendered dataset based on Unreal Engine with the following two key designs:
(1) \textbf{Customized scenes and dynamic objects.} We curate a diverse collection of scenes with occlusion relationships, including factories, streets, and natural landscapes, as well as dynamic objects such as humans and animals with various actions. By combining different scenes and dynamic objects, we construct diverse dynamic scenarios.
(2) \textbf{Automated trajectory generation.} To enable automated trajectory generation with collision avoidance in dynamic scenes, we implement a blueprint script in Unreal Engine. This script can randomly generate camera trajectories of arbitrary length with automatic obstacle avoidance, and the traversal range is customizable. This significantly improves our data collection efficiency.
Using this pipeline, we collect $100$ long videos across $13$ scenes, with each video averaging over $18000$ frames, totaling $16.7$ hours of video content. Additional details about data collection can be found in Appendix~\ref{sec:supp_dataset}.

\subsection{Multi-Dataset Training Strategy}
\label{subsec:training_strategy}
\input{fig_tab_tex/tab_dataset_comp}
Video datasets can be categorized into three types based on camera pose annotation accuracy and visual realism: (1) Rendered data~\cite{cam} with precise pose annotations but non-photorealistic style and limited diversity; (2) Real-world data with estimated poses~\cite{spatialvid}, offering photorealistic style but less precise annotations and limited dynamics due to the filtering required for accurate pose estimation; and (3) Real-world data without accurate pose annotations~\cite{sekai}, providing photorealistic style, rich content diversity, and better dynamics, but lacking accurate camera annotations.


To leverage the complementary strengths of these data sources, we propose a multi-dataset training strategy that assigns a dedicated camera encoder to each dataset type. Specifically, rendered data with precise poses and estimated-pose data each use a separate camera encoder to process their respective camera annotations, while data without reliable poses is fed with zero camera parameters (i.e., $\mathbf{R}=\mathbf{0},\mathbf{t}=\mathbf{0}$) through a third dedicated camera encoder. By isolating different annotation qualities into separate encoders, datasets with varying pose accuracy do not interfere with each other during training, allowing the model to benefit from precise pose supervision of rendered data, the realism of estimated-pose data, and the diversity of unannotated real-world data simultaneously. During inference, we use only the camera encoder trained on precisely annotated data to ensure reliable camera control.

%% file: fig_tab_tex/dit.tex
\begin{figure}[t]
  \centering
  \includegraphics[width=0.7\linewidth]{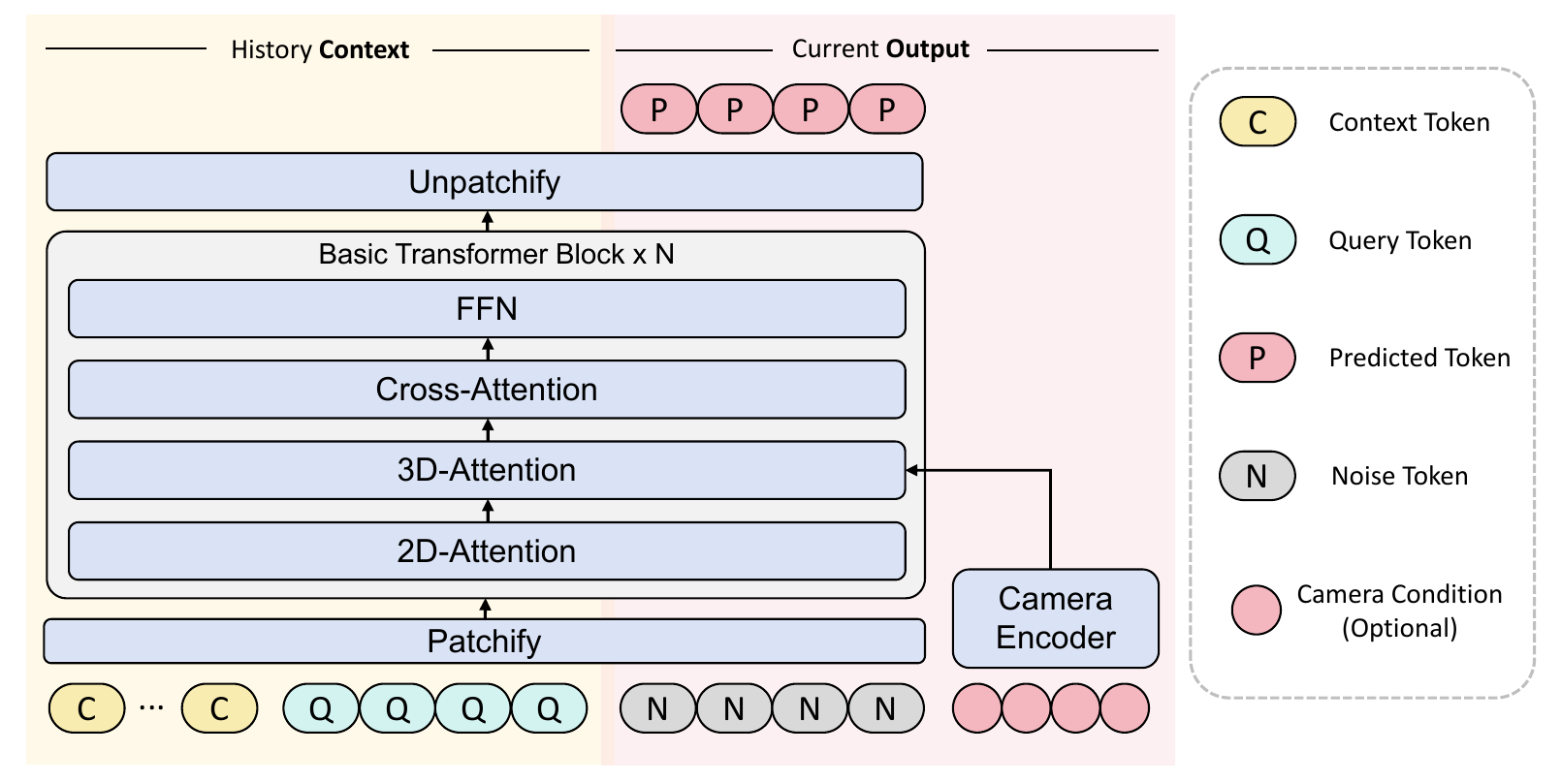}
\caption{\textbf{Model Architecture.} Our video generation model adopts a Diffusion Transformer. We concatenate $\mathbf{C}$, $\mathbf{Q}$, $\mathbf{P}$ tokens for context-conditioned long video generation. An optional camera encoder supports interactive control.}
  \vspace{-0.5cm}
\label{fig:dit} 
\end{figure}

%% file: fig_tab_tex/framework.tex
\begin{figure}[t]
  \centering
  \includegraphics[width=1\linewidth]{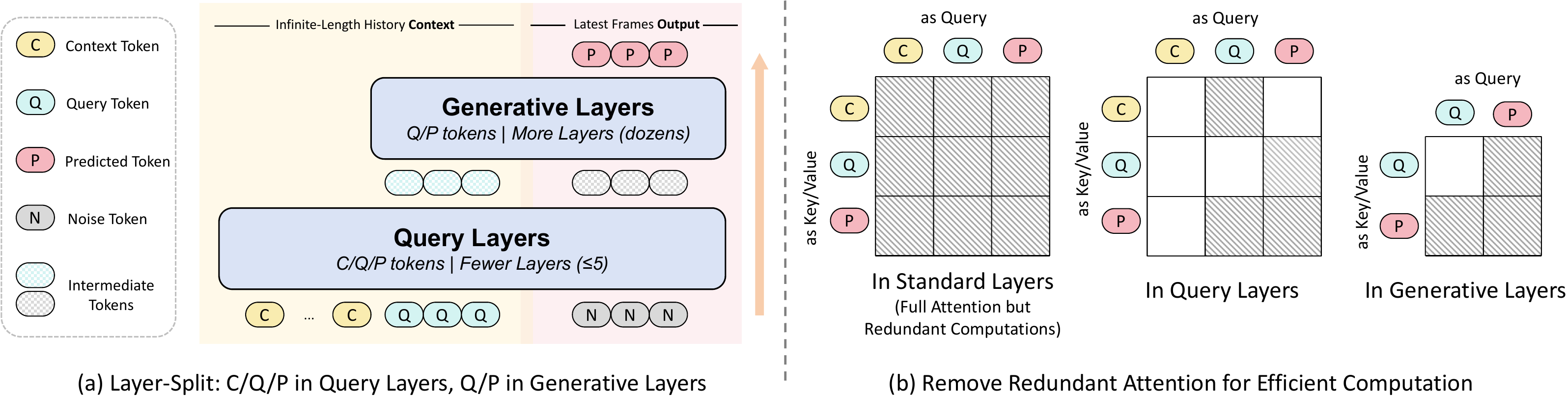}
  \caption{\textbf{Efficiency Strategies.} (a) Strategy\#1: Context querying in shallow Query Layers with $\mathbf{C}$, $\mathbf{Q}$, $\mathbf{P}$ tokens; deep Generative Layers use only $\mathbf{Q}$, $\mathbf{P}$ tokens. (b) Strategy\#2: Remove unnecessary attention computation for improved efficiency.}
  \vspace{-0.5cm}
\label{fig:framework} 
\end{figure}

%% file: fig_tab_tex/tab_dataset_comp.tex
\begin{table}[t]
\centering

\caption{\textbf{Comparison of long video datasets for video world models.} No existing dataset fully satisfies all four requirements: precise per-frame camera pose annotations, occlusion relationships, dynamic objects, and revisit scenarios. We collect a dataset meeting these requirements.
\textsuperscript{1}HQ subset only.
\textsuperscript{2}Available subset only.
\textsuperscript{3}Insufficient annotation accuracy.
\textsuperscript{4}Not all frames are annotated.
\textsuperscript{5}Primarily street walking videos with limited occlusions.
\textsuperscript{6}Filtered videos with reduced dynamics.
\textsuperscript{7}Few revisits without specialized design, especially for real-world data.
}

\newcommand{\gou}{\textcolor{ForestGreen}{\ding{52}}}
\newcommand{\med}{\textcolor{orange}{\ding{108}}}  
\newcommand{\cha}{\textcolor{Red}{\ding{55}}}

\vspace{-0.2cm}
\small{\gou: Fully satisfied \quad \med: Partially satisfied \quad \cha: Not satisfied}
\vspace{0.1cm}

\fontsize{8}{9}\selectfont
\begin{tabular}{l|c|c|c|c|c|c|c}
\toprule
Dataset & Source & Style &  Duration & Cam. Pose & Occlusion & Dynamic & Revisit \\
\hline\hline
CaM~\cite{cam} & Simulator & Rendered & $7.0$ h & \gou & \cha & \cha & \gou  \\
SpatialVid~\cite{spatialvid} & YouTube & Real-World & $1146.06$ h\textsuperscript{1} & \gou & \med\textsuperscript{5} & \med\textsuperscript{6} & \cha\textsuperscript{7} \\
Sekai-real~\cite{sekai} & YouTube & Real-World & $304.0$ h\textsuperscript{1} & \med\textsuperscript{3} & \med\textsuperscript{5} & \gou & \cha\textsuperscript{7} \\
OmniWorld~\cite{omniworld} & Simulator & Rendered  & $11.14$ h\textsuperscript{2} & \cha\textsuperscript{4} & \gou & \gou & \med\textsuperscript{7}   \\
Ours & Simulator & Rendered  & $16.7$ h & \gou & \gou & \gou & \gou  \\
\bottomrule
\end{tabular}\\

\vspace{-0.3cm}
\label{tab:dataset_comparison}
\end{table}

%% file: sec/4_exp.tex
\section{Experiments}
\label{sec:exp}



\subsection{Implementation Details.}
\label{subsec:impl}
In our implementation, $\mathbf{P}$ tokens correspond to $77$ video frames per generation. The number of $\mathbf{Q}$ tokens equals that of $\mathbf{P}$ tokens. Rather than using learnable parameters, $\mathbf{Q}$ tokens are initialized as randomly sampled noise, which better aligns with the input distribution of diffusion models. During training and inference, $\mathbf{C}$ tokens can be up to $9$ times the length of $\mathbf{P}$ tokens, with the last frame of the $\mathbf{C}$ token sequence always matching the first frame of the $\mathbf{P}$ token sequence to ensure continuity. During training, the length of $\mathbf{C}$ tokens is uniformly sampled from $0$ to $9$ times the $\mathbf{P}$ token length, where $0$ corresponds to training only the image-to-video capability. We perform full fine-tuning of the video generation model on the collected dataset.

Our primary experiments are built upon an internal 1B-parameter text-to-video diffusion transformer with $28$ layers, of which $5$ serve as Query Layers. We select this model for its stronger visual quality than open-source models of comparable scale and far lower computational costs than larger alternatives (\eg, 5B, 14B), enabling more comprehensive ablation studies with higher-quality baselines.
To verify that \textsc{MemLearner} generalizes across architectures, we additionally apply our method to Wan2.1 (T2V-1.3B)~\cite{wan}, an open-source video DiT. As detailed in Appendix~\ref{sec:open_source}, \textsc{MemLearner} achieves consistent improvements over the CaM~\cite{cam} baseline on the open-source model, confirming that our approach is architecture-agnostic and benefits from pre-trained priors regardless of the specific backbone.

Videos are generated at $640\times 352$ resolution, producing $77$ frames with a causal 3D VAE that applies a temporal compression ratio of $4$, yielding $20$-frame video latents. We configure $\mathbf{P}$ tokens to represent $20$ video latent frames, with $\mathbf{Q}$ tokens matching the $\mathbf{P}$ token count, while $\mathbf{C}$ tokens can represent up to $180$ video latent frames. Training is conducted for over $20{,}000$ iterations with a batch size of $8$ and a learning rate of $5\times10^{-5}$. We explore two dataset configurations: (1)~\textbf{Rendered-only}: $50\%$ from our collected dataset and $50\%$ from CaM~\cite{cam}; (2)~\textbf{Mixed}: $75\%$ rendered data (combining configuration~1) and $25\%$ real-world data ($12.5\%$ Sekai-real~\cite{sekai} and $12.5\%$ SpatialVid~\cite{spatialvid}). For sampling, we apply Classifier-Free Guidance~\cite{cfg} with $50$ steps for text conditioning.



\subsection{Evaluation Methods}
For evaluation, we reserve a 5\% held-out test set via random dataset partitioning with no video overlap between training and test splits.
Our evaluation protocol includes: (1) \textbf{FID/FVD} to assess video visual quality; (2) \textbf{PSNR/LPIPS} to measure memory capability by quantifying pixel-wise differences between frames. Following prior work~\cite{cam}, we adopt two evaluation strategies for memory assessment: (1) \textbf{Ground truth comparison (GT Comp.)}: measuring whether predicted frames align with ground truth given context from ground truth frames. With sufficient context, generated results should match the ground truth; (2) \textbf{Revisit comparison (Revisit Comp.)}: comparing newly generated revisit frames against previously generated ones in long video sequences. For implementation, we test on simple camera trajectories where the camera rotates $n$ degrees with $n \sim \mathcal{U}(90^\circ, 180^\circ)$ and returns, enabling straightforward identification of corresponding frame pairs. We then compute PSNR/LPIPS between all corresponding frame pairs from the outbound and return paths and average the results, thereby evaluating consistency throughout the entire generation process rather than just between the first and last frames.

To ensure evaluation diversity, we conduct experiments across multiple datasets with distinct characteristics: our collected dataset with occlusions and dynamic objects (Tab.~\ref{tab:comp}, Fig.~\ref{fig:comparison}), the CaM dataset~\cite{cam} without occlusions or dynamics (Tab.~\ref{table:comp_cam_dataset}), and the real-world SpatialVID dataset~\cite{spatialvid} (Appendix~\ref{sec:spatialvid}), where \textsc{MemLearner} consistently outperforms baselines across all three evaluation settings. We further provide a user study in Appendix~\ref{sec:user_study}, in which \textsc{MemLearner} is preferred over other SOTAs in terms of visual quality and scene consistency.


\subsection{Qualitative Results}
\label{subsec:qualitative}
\input{fig_tab_tex/fig_qualitative_render}
\input{fig_tab_tex/fig_qualitative_real}
We present qualitative examples to illustrate \textsc{MemLearner} generation quality. As shown in Fig.~\ref{fig:qualitative_rendered}, our method handles diverse indoor/outdoor scenes with rich occlusions and dynamic objects. We provide video demos on our \href{https://yujiwen.github.io/memlearner/}{project page} for direct viewing. Since our mixed training uses real-world data, \textsc{MemLearner} generalizes to real-world domains without adaptation. Fig.~\ref{fig:qualitative_real} shows the model preserves scene layout and object appearance across revisits, confirming the learned query mechanism generalizes to real environments.

\subsection{Comparison Results}
\label{subsec:exp_comp}
\input{fig_tab_tex/fig_comp}
\input{fig_tab_tex/tab_comp}

In this section, we compare baselines and SOTAs:  
(1) DFoT~\cite{dfot}, a long video generation model without memory design;  
(2) FramePack~\cite{framepack}, which hierarchically compresses history frames as conditions to retain limited memory;  
(3) VMem~\cite{vmem}, which achieves memory capability via a point-cloud-based retrieval rule for context retrieval.
(4) Context-as-Memory~\cite{cam}, which selects relevant conditioning frames via FOV-overlap computation to provide effective memory;
(5) Separate Module, i.e., the design in Fig.~\ref{fig:arch_clarify} (b), where an additional five-layer Transformer (with the same layer structure as the video DiT) serves as a context query module. Its output channels are aligned with the video DiT's internal features, eliminating the need for a patchify layer;  
(6) our proposed \textsc{MemLearner}.

For fair comparison, all methods are implemented using our codebase and dataset. We evaluate both memory-related metrics (PSNR/LPIPS) and visual quality metrics (FID/FVD). Results are summarized in Tab.~\ref{tab:comp} and Fig.~\ref{fig:comparison}. \textsc{MemLearner} achieves the best performance across all metrics and benchmarks. 
Separate Module achieves poor performance on GT Comp., as it fails to learn conditioning on context information and behaves like a text-to-video model. This demonstrates that jointly training a from-scratch context query module with a pre-trained video generation model cannot effectively learn context query capability; instead, the video generation model ignores the context query module. In contrast, \textsc{MemLearner}, which leverages the video generation model itself for context querying, is significantly more effective.

In addition to our collected dataset, we also conduct quantitative evaluations on CaM dataset~\cite{cam}, which lacks occlusion and dynamic object scenes as shown in Tab.~\ref{tab:dataset_comparison}. Results in Tab.~\ref{tab:comp} and Tab.~\ref{table:comp_cam_dataset} show CaM and our method perform comparably on CaM dataset (no occlusions/dynamics), while CaM degrades sharply on our dataset and our method remains robust. This directly validates our approach’s effectiveness for handling occlusions and dynamic objects.

\subsection{Ablation Study}
\label{subsec:ablation}
\paragraph{Ablation of Attention Computation.}
\input{fig_tab_tex/tab+fig_cam-dataset-comp+attention-comp}
\input{fig_tab_tex/tab_ablation_attn+cam}
As shown in Fig.~\ref{fig:attention_ablation}, we compare three attention variants: (a) standard setting; (b) augmenting (a) with $\mathbf{Q}$ tokens as queries attending to $\mathbf{Q}$ tokens as keys/values; (c) removing from (a) the pattern where $\mathbf{Q}$ tokens as queries attend to $\mathbf{P}$ tokens as keys/values. Results in Tab.~\ref{table:ablation_attention} show that (b) matches (a) in performance, the added computation is non-critical and removable. In contrast, (c) exhibits significant performance degradation, demonstrating that it is essential for $\mathbf{Q}$ tokens to extract guidance from $\mathbf{P}$ tokens to determine what information to query from $\mathbf{C}$ tokens.
We further analyze the effect of including $\mathbf{C}$ tokens as queries in Appendix~\ref{sec:attn_compute}, which introduces substantial computational overhead with negligible benefit.

\paragraph{Ablation of Query Layer Number.}
We ablate Query Layer count impact in Tab.~\ref{table:ablation_layer}. Performance saturates at 5 layers, with computational overhead rising thereafter. 5 layers balance computational cost and memory performance.

\paragraph{Ablation of Camera Embedding.}
\input{fig_tab_tex/tab_ablation_layer+dataset}
When injecting camera pose embeddings, providing them to $\mathbf{P}$ tokens is essential for interactive video generation. However, it is unclear whether $\mathbf{C}$ tokens and $\mathbf{Q}$ tokens also require camera pose information. We investigate this in Tab.~\ref{table:ablation_cam}, comparing two settings: injecting camera poses only into $\mathbf{P}$ tokens versus all tokens. For $\mathbf{C}$ tokens, we inject the corresponding camera pose, while $\mathbf{Q}$ tokens receive the same as $\mathbf{P}$ tokens. Surprisingly, the results show that even without providing camera poses to $\mathbf{C}$ and $\mathbf{Q}$ tokens, the memory performance does not degrade significantly. This suggests that video generation models have the potential to learn geometric correspondences implicitly without explicit 3D information such as camera poses.

\paragraph{Ablation of Datasets.}
In Tab.~\ref{table:comp_mixed_dataset}, we ablate the impact of training with different dataset mixtures. `CaM' refers to the dataset proposed in Context-as-Memory~\cite{cam}, `Real' denotes the mixture of SpatialVID~\cite{spatialvid} and Sekai-real~\cite{sekai} real-world datasets, and `Ours' is our proposed dataset. When combining rendered datasets `CaM' and `Ours', we use a 1:1 ratio, while rendered and real datasets are mixed at a 3:1 ratio. The `Real Quality' metric measures the FID/FVD distance between generated videos and real-world videos. We observe that training solely on real-world data fails to learn memory capabilities, as evidenced by poor performance across all metrics, indicating insufficient revisit patterns in these real-world datasets. Models trained purely on rendered data produce videos whose distribution deviates from real videos. 
Incorporating real data into training effectively improves the realism of generated videos, as further illustrated by qualitative comparisons in Appendix~\ref{sec:dataset_ablation}.



%% file: fig_tab_tex/fig_qualitative_render.tex
\begin{figure}[t]
  \centering
  \includegraphics[width=1\linewidth]{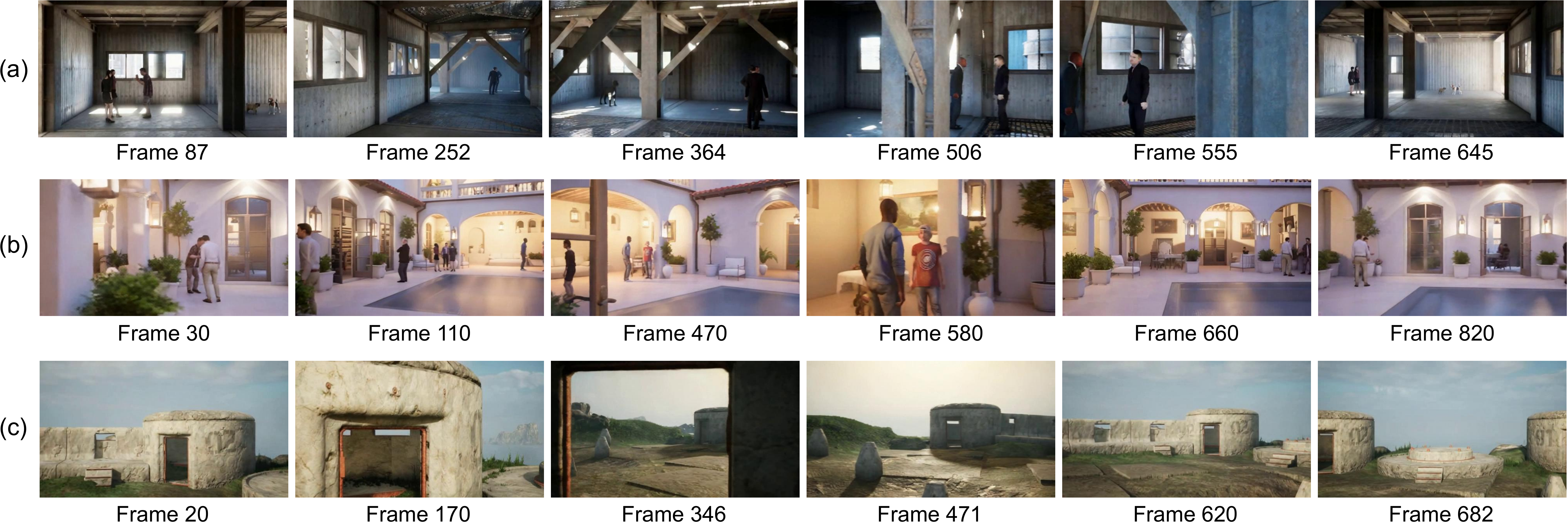}
  \vspace{-0.5cm}
  \caption{\textbf{Qualitative Results.} \textsc{MemLearner} effectively handles both indoor (a, b) and outdoor (c, Fig.~\ref{fig:teaser}) scenes with occlusions and dynamic objects.}
  \vspace{-0.2cm}
\label{fig:qualitative_rendered} 
\end{figure}

%% file: fig_tab_tex/fig_qualitative_real.tex
\begin{figure}[t]
  \centering
  \includegraphics[width=1\linewidth]{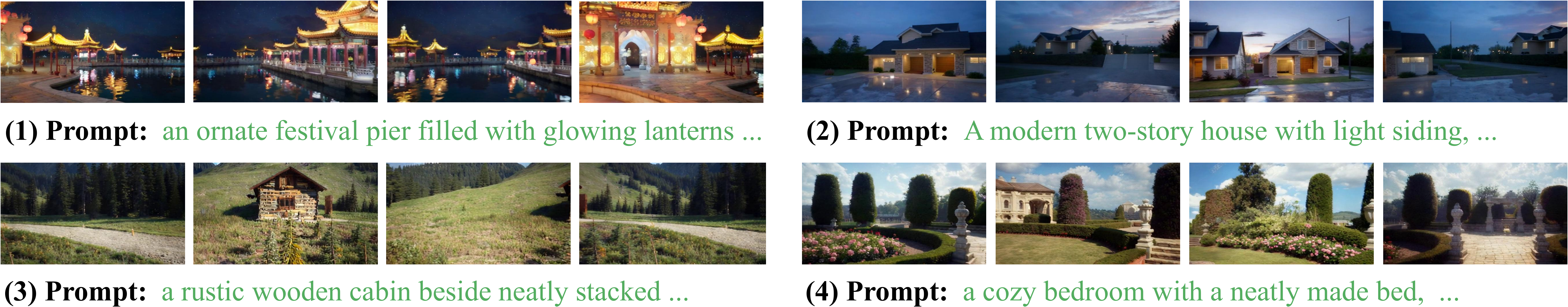}
  \vspace{-0.5cm}
  \caption{\textbf{Real-World Qualitative Results.} By incorporating real-world videos into training dataset, \textsc{MemLearner} generalizes to real-world scenes.}
  \vspace{-0.2cm}
\label{fig:qualitative_real} 
\end{figure}

%% file: fig_tab_tex/fig_comp.tex
\begin{figure}[t]
  \centering
  \includegraphics[width=1\linewidth]{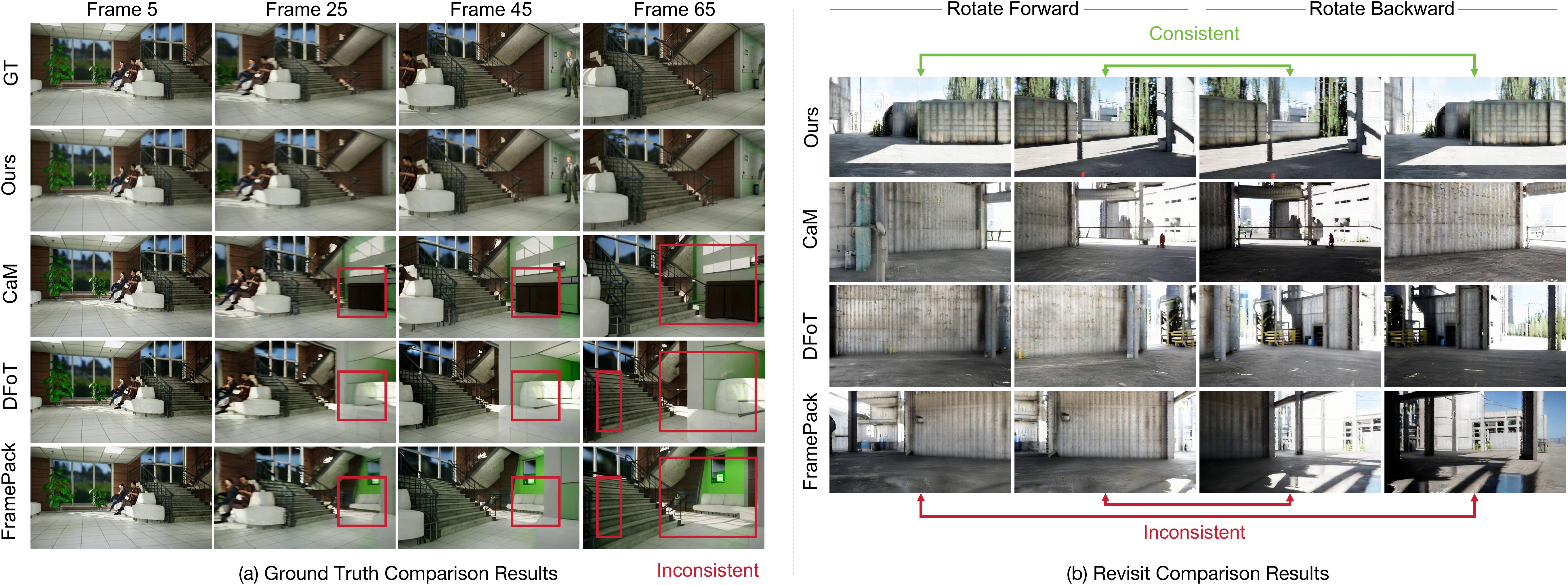}
  \vspace{-0.3cm}
  \caption{\textbf{Qualitative Comparison.} \textsc{MemLearner} achieves optimal memory and visual quality, demonstrating learning-based adaptive context query effectiveness. Other methods show inconsistent memory performance.}
  \vspace{-0.2cm}
\label{fig:comparison} 
\end{figure}

%% file: fig_tab_tex/tab_comp.tex
\begin{table}[t]

\tabcolsep=0.09cm
\centering

\caption{\textbf{Quantitative Comparison.} \textsc{MemLearner} outperforms all metrics. Notably, the Separate Module (Fig.~\ref{fig:arch_clarify} (b)) lacks context learning capability, indicating joint training of a scratch module with pre-trained video DiT is ineffective.}

\vspace{-0.2cm}
\fontsize{8}{9}\selectfont
\begin{tabular}{l | c c c c | c c c c | c }
\toprule
& \multicolumn{4}{c}{GT Comp.} & \multicolumn{4}{c}{Revisit Comp.} & \\
  Methods  & PSNR$\uparrow$ & LPIPS$\downarrow$ & FID$\downarrow$ & FVD$\downarrow$ & PSNR$\uparrow$ & LPIPS$\downarrow$ & FID$\downarrow$ & FVD$\downarrow$ & fps$\uparrow$  \\
  \hline
  \hline

  DFoT~\cite{dfot} & 16.98 & 0.4796 & 147.09 & 998.43 & 16.14 & 0.5481 & 151.38 & 1021.46 & \textbf{1.59} \\
  FP~\cite{framepack} & 16.42 & 0.5104 & 143.97 & 967.97 & 15.86 & 0.5837 & 154.11 & 1037.58 & 1.40 \\
  VMem~\cite{vmem} & 19.59 & 0.3872 & 129.94 & 850.17 & 17.30 & 0.4187 & 141.82 & 968.45 & 0.73 \\
  CaM~\cite{cam} & 19.85 & 0.3475 & 125.35 & 848.61 & 17.61 & 0.3934 & 137.87 & 948.63 & 0.97 \\
Fig.~\ref{fig:arch_clarify} (b) & 9.16 & 0.6567 & 145.63 & 930.54 & - & - & - & - & 0.48 \\
Ours & \textbf{21.23} & \textbf{0.2904} & \textbf{112.75} & \textbf{835.98} & \textbf{18.57} & \textbf{0.3230} & \textbf{101.57} & \textbf{847.52} & 0.54 \\

\bottomrule
\end{tabular}


\vspace{-0.3cm}
\label{tab:comp}
\end{table}

%% file: fig_tab_tex/tab+fig_cam-dataset-comp+attention-comp.tex
\begin{table}[t]
\centering
\tabcolsep=0.06cm

\begin{minipage}{0.45\textwidth}
\centering
\fontsize{6}{7}\selectfont
\caption{Quantitative comparison on CaM dataset~\cite{cam} (no occlusions/dynamics). Small performance gaps, widening sharply on our occlusion/dynamic dataset (Tab.~\ref{tab:comp}).}
\vspace{-0.2cm}
\begin{tabular}{l | c c | c c}
\toprule
& \multicolumn{2}{c}{GT Comp.} & \multicolumn{2}{c}{Revisit Comp.}\\
Methods  & PSNR$\uparrow$ & LPIPS$\downarrow$ & PSNR$\uparrow$ & LPIPS$\downarrow$ \\
\hline\hline
VMem~\cite{vmem} & 19.84 & 0.3564 & 17.98 & 0.3633 \\
CaM~\cite{cam} & 20.22 & 0.3003 & 18.11 & 0.3414 \\
Ours & \textbf{20.35} & \textbf{0.2975} & \textbf{18.29} & \textbf{0.3374} \\
\bottomrule
\end{tabular}
\label{table:comp_cam_dataset}
\end{minipage}
\hfill
\begin{minipage}{0.48\textwidth}
\centering
\includegraphics[width=\linewidth]{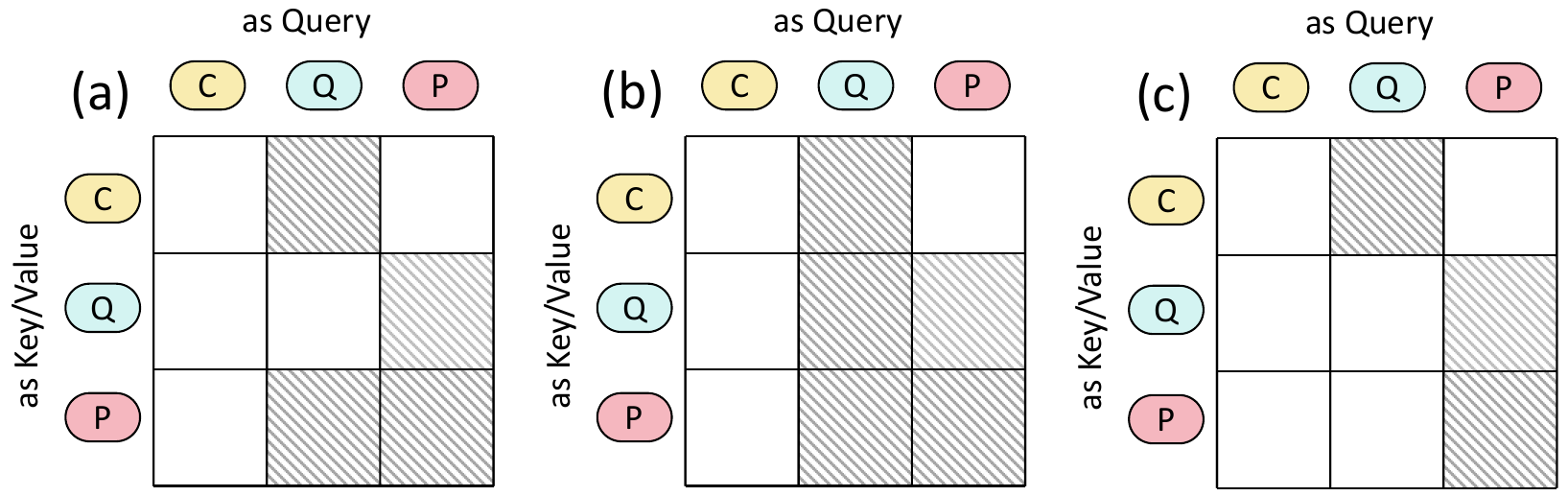}
\vspace{0.2cm}
\captionof{figure}{Attention computation settings for ablation study in Tab.~\ref{table:ablation_attention}. Performance impact of attention scope when \textbf{Q} tokens act as queries.}
\label{fig:attention_ablation}
\end{minipage}

\vspace{-0.5cm}

\end{table}

%% file: fig_tab_tex/tab_ablation_attn+cam.tex
\begin{table}[t]
\centering
\tabcolsep=0.06cm 

\begin{minipage}{0.48\textwidth}
\centering

\caption{Ablation of Attention Computation: Settings (a)-(c) in Fig.~\ref{fig:attention_ablation}. Critical computation is \textbf{Q} queries attending to \textbf{P} keys/values.}

\fontsize{6}{7}\selectfont 
\begin{tabular}{c | c c | c c | c}
\toprule
& \multicolumn{2}{c}{GT Comp.} & \multicolumn{2}{c}{Revisit Comp.} &\\
Setting  & PSNR$\uparrow$ & LPIPS$\downarrow$ & PSNR$\uparrow$ & LPIPS$\downarrow$  & fps$\uparrow$ \\
\hline\hline
(a) & \textbf{21.23} & \textbf{0.2904} & \textbf{18.57} & \textbf{0.3230} & 0.54 \\
(b) & 21.19 & 0.2949 & 18.53 & 0.3342 & 0.51 \\
(c) & 17.27 & 0.4657 & 16.34 & 0.5107 & \textbf{0.56} \\
\bottomrule
\end{tabular}

\label{table:ablation_attention}
\end{minipage}
\hfill
\begin{minipage}{0.48\textwidth}
\centering

\caption{Ablation of camera embedding. Even without camera poses for $\mathbf{C}$ tokens, the model learns to query useful information with no significant performance drop.}

\fontsize{6}{7}\selectfont 
\begin{tabular}{c | c c | c c }
\toprule
& \multicolumn{2}{c}{GT Comp.} & \multicolumn{2}{c}{Revisit Comp.} \\
Add Camera Emb.  & PSNR$\uparrow$ & LPIPS$\downarrow$ & PSNR$\uparrow$ & LPIPS$\downarrow$   \\
\hline\hline
All tokens & \textbf{21.23} & \textbf{0.2904} & \textbf{18.57} & \textbf{0.3230} \\
Only P tokens & 21.17 & 0.2946 & 18.38 & 0.3384 \\
\bottomrule
\end{tabular}

\label{table:ablation_cam}
\end{minipage}

\vspace{-0.5cm}

\end{table}

%% file: fig_tab_tex/tab_ablation_layer+dataset.tex
\begin{table}[t]
\centering
\tabcolsep=0.06cm  

\begin{minipage}{0.48\textwidth}
\centering

\caption{Ablation of Query Layer Count: 5 layers balance computational cost and memory performance.}

\fontsize{6}{7}\selectfont  
\begin{tabular}{c | c c | c c | c}
\toprule
& \multicolumn{2}{c}{GT Comp.} & \multicolumn{2}{c}{Revisit Comp.} &\\
\#Layer  & PSNR$\uparrow$ & LPIPS$\downarrow$ & PSNR$\uparrow$ & LPIPS$\downarrow$  & fps$\uparrow$ \\
\hline\hline
1  & 18.16 & 0.4056 & 17.25 & 0.4144 & \textbf{0.61} \\
3  & 19.43 & 0.3625 & 17.83 & 0.3808 & 0.58 \\
5  & 21.23 & 0.2904 & 18.57 & 0.3230 & 0.54 \\
10 & 21.36 & 0.2913 & 18.50 & \textbf{0.3214} & 0.46 \\
20 & \textbf{21.37} & \textbf{0.2891} & \textbf{18.66} & 0.3217 & 0.36 \\
\bottomrule
\end{tabular}

\label{table:ablation_layer}
\end{minipage}
\hfill
\begin{minipage}{0.48\textwidth}
\centering

\caption{Mixed Dataset Training Ablation: Real-world data alone cannot learn memory capability (insufficient revisit).}

\fontsize{6}{7}\selectfont  
\begin{tabular}{l | c c | c c}
\toprule
& \multicolumn{2}{c}{Revisit Comp.} & \multicolumn{2}{c}{Real Quality} \\
Datasets  & PSNR$\uparrow$ & LPIPS$\downarrow$ & FID$\downarrow$ & FVD$\downarrow$ \\
\hline\hline
CaM~\cite{cam} & 18.35 & 0.3263 & 167.47 & 1368.01 \\
Real~\cite{sekai, spatialvid} & 15.32 & 0.6019 & \textbf{104.16} & \textbf{828.95} \\
CaM+Real & 18.27 & 0.3312 & 111.33 & 849.98 \\
CaM+Ours & \textbf{18.57} & \textbf{0.3230} & 161.83 & 1379.42 \\
CaM+Ours+Real & 18.49 & 0.3251 & 115.47 & 837.91 \\
\bottomrule
\end{tabular}

\label{table:comp_mixed_dataset}
\end{minipage}

\vspace{-0.5cm}

\end{table}

%% file: sec/5_conclusion.tex
\section{Conclusion}
\label{sec:conclusion}
In this work, we propose a paradigm shift for context memory in video world models, transitioning from rule-based context retrieval to learning-based context query. To validate our method, we collect rendered data with accurate camera annotations, customized occlusions and dynamic objects, and sufficient revisit patterns. We further propose a multi-dataset training strategy to effectively leverage both rendered and real-world videos.
\paragraph{Limitations and Future Work.}
While \textsc{MemLearner} advances context memory in video world models, the memory problem remains far from solved:
(1) Expanding model capacity and training data scale is imperative. At our current 1B model scale, scenes with many simultaneously interacting characters in large environments remain challenging, as the model lacks sufficient capacity to track numerous dynamic entities. We observe that generation quality degrades notably when more than five characters interact within the same scene, producing inconsistent appearances or missing objects upon revisit. Scaling to larger models and collecting long video data with richer dynamic interactions are necessary to address this.
(2) Current research, including ours, focuses on full context storage and efficient information retrieval/querying~\cite{cam,lct,far,moc,worldmem,vmem,ssvwm}, yet memory should not scale linearly with generation time. Notably, context compression is orthogonal to our contribution: one can first compress context representations and then apply our learned query mechanism on the compressed tokens. Exploring compression, summarization, updating, editing and selective forgetting are key future directions for practical, intelligent memory systems.

%% file: tex/dataset.tex
\section{Details of Collected Dataset}
\label{sec:supp_dataset}

\paragraph{3D Scenes and Dynamic Objects.}
We collect 13 diverse 3D scene assets from Fab.com\footnote{\url{https://www.fab.com/}}. To minimize the domain gap between rendered data and real-world videos, we prioritize photorealistic scene assets. We also incorporate stylized scenes to enhance dataset diversity. The collected scenes encompass various environments including streets, shopping malls, rural areas, indoor and outdoor spaces. To increase scene dynamics, we introduce dynamic objects including human characters with diverse appearances and animals such as dogs, camels, and horses. By randomly combining these scenes with dynamic objects, we construct diverse dynamic 3D scenarios for training data generation.

\paragraph{Automated Camera Trajectory Generation.}
To automate camera trajectory generation, we implement a Blueprint script in Unreal Engine. The script introduces a camera actor with randomized movement controlled by the script logic. In play mode, this actor autonomously navigates the scene using Unreal Engine's built-in navigation APIs for obstacle avoidance and random exploration. The script logs the camera pose at each timestep to a file. Subsequently, we parse this log file to generate sequence files compatible with Unreal Engine's renderer, enabling direct video rendering. For rendering, the camera is configured with a focal length of 24mm, an aperture of 10, and a field of view (FOV) of 52.67 degrees.

\paragraph{Data Preprocessing and Text Annotation.}
The rendered video frames are captioned using a pre-trained multimodal large language model~\cite{minicpm}. We generate captions every 77 frames. During training, we randomly sample 77 frames from a long video sequence and use the caption of the nearest annotated segment as the text condition for training.

%% file: tex/model.tex
\section{Details of Internal Model Architecture}
\label{sec:supp_model}

\paragraph{Base Text-to-Video Generation Model}
Our base T2V generation model adopts a latent diffusion transformer architecture, shown in Figure~\ref{fig_basemodel}. Videos are first encoded into latent representations via a 3D-VAE, which then serve as the input to a diffusion transformer. Previous approaches using UNets or transformers often append a separate 1D temporal attention module for temporal modeling. However, this spatially-temporally decoupled design is suboptimal. We adopt 3D self-attention to directly model spatiotemporal tokens, leading to improved coherence and quality in generated videos. We derive scale parameters from the diffusion timestep and apply RMSNorm to spatiotemporal tokens before each attention and feed-forward network (FFN) layer.
\input{subtex-fig-tex/fig_dit_arch}

\paragraph{3D Attention Implementation Details.}
Our designed 3D attention efficiently processes three types of tokens: context, query, and predicted tokens. We provide the corresponding pseudocode in Figure~\ref{code:3d_attention}.
\input{subtex-fig-tex/fig_pseudocode}

%% file: subtex-fig-tex/fig_dit_arch.tex
\begin{figure*}[h]\centering
\includegraphics[width=0.7\textwidth]{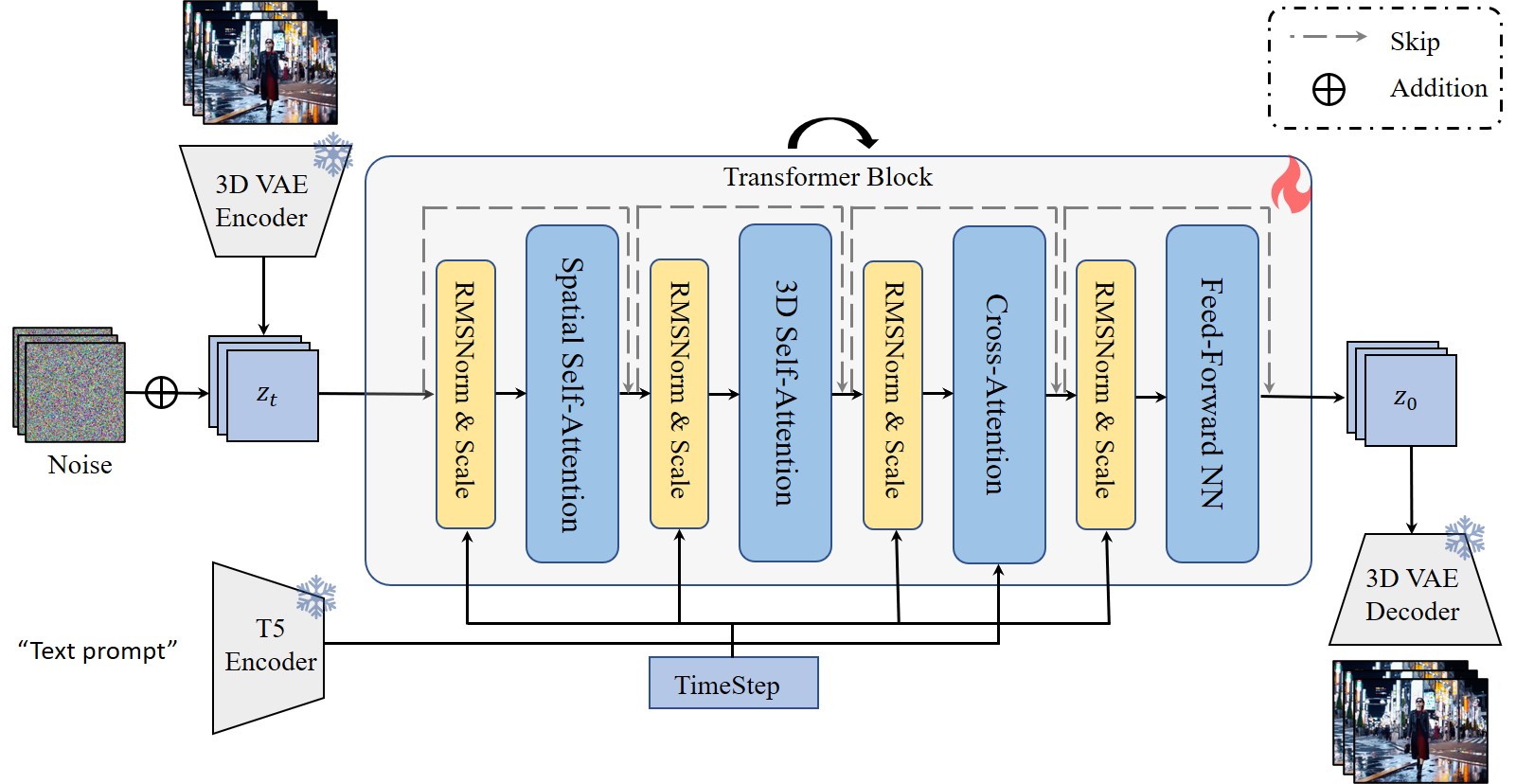}
\caption{\text{Overview of the base text-to-video generation model.}}
    \label{fig_basemodel}
\end{figure*}

%% file: subtex-fig-tex/fig_pseudocode.tex
\begin{figure}[t]
\begin{lstlisting}[language=Python]
# Extract tokens from hidden states
def extract_tokens(hs):
    """Split hs into P, Q, C tokens based on frame indices"""
    return P_tokens, Q_tokens, C_tokens

# 3D Attention Computation
if layer_type == "Generative":
    # Extract P and Q tokens only
    Q_query, P_query = extract_tokens(query)
    Q_key, P_key = extract_tokens(key)
    Q_value, P_value = extract_tokens(value)

    # Q attends to P; P attends to P+Q
    Q_out = flash_attention(Q_query, P_key, P_value)
    P_out = flash_attention(P_query, concat(P_key, Q_key), concat(P_value, Q_value))

elif layer_type == "Query":
    # Extract C, Q, P tokens
    Q_query, P_query, C_query = extract_tokens(query)
    Q_key, P_key, C_key = extract_tokens(key)
    Q_value, P_value, C_value = extract_tokens(value)

    # Q attends to P+C; P attends to P+Q
    Q_out = flash_attention(Q_query, concat(P_key, C_key), concat(P_value, C_value))
    P_out = flash_attention(P_query, concat(P_key, Q_key), concat(P_value, Q_value))

# Concatenate and project output
output = concat(P_out, Q_out)
hidden_states = output_projection(output)
\end{lstlisting}
\vspace{-0.5cm}
\caption{Pseudocode for 3D attention computation. Generative Layers process P and Q tokens, while Query Layers process all C, Q, P tokens with different attention patterns.}
\label{code:3d_attention}
\end{figure}

%% file: tex/exp.tex
\newpage
\section{Supplementary Experimental Results}
\label{sec:supp_exp}

\subsection{Results on Open-Source Video Models}
\label{sec:open_source}

To verify that MemLearner generalizes across different models, we apply our method to Wan~2.1 (T2V-1.3B)~\cite{wan}, an open-source video Diffusion Transformer~\cite{dit, sora}. Wan~2.1 shares a similar architecture with our internal model: each Transformer block contains a 3D attention module, making it directly compatible with our context query mechanism. We follow the same training protocol as described in Sec.~\ref{subsec:impl} of the main text and compare against CaM~\cite{cam} and VMem~\cite{vmem}.

As shown in Tab.~\ref{table:comp_wan}, MemLearner consistently outperforms both baselines on Wan~2.1, confirming that our learning-based context query approach is architecture-agnostic. We note that the absolute performance of all methods on Wan~2.1 is lower than on our internal model (Tab.~\ref{tab:comp} in the main text), which we attribute to the relatively weaker generation capability of this open-source model at the 1.3B scale.

\begin{table}[h]
\centering
\begin{tabular}{l | c c | c c}
\toprule
& \multicolumn{2}{c|}{GT Comp.} & \multicolumn{2}{c}{Revisit Comp.}\\
Methods  & PSNR$\uparrow$ & LPIPS$\downarrow$ & PSNR$\uparrow$ & LPIPS$\downarrow$ \\
\hline\hline
VMem~\cite{vmem} & 18.24 & 0.4282 & 16.87 & 0.4738  \\
CaM~\cite{cam} & 18.37 & 0.4192 & 16.91 & 0.4640 \\
Ours & \textbf{19.25} & \textbf{0.3913} & \textbf{17.17} & \textbf{0.4472} \\
\bottomrule
\end{tabular}
\caption{Quantitative comparison on Wan~2.1 (T2V-1.3B)~\cite{wan}.}
\label{table:comp_wan}
\end{table}

\subsection{Attention Visualization Results}
\label{sec:attn_vis}

We visualize the attention similarity between query tokens and context tokens to validate two key claims in the main text.

\begin{figure}[h]
  \centering
  \includegraphics[width=0.49\linewidth]{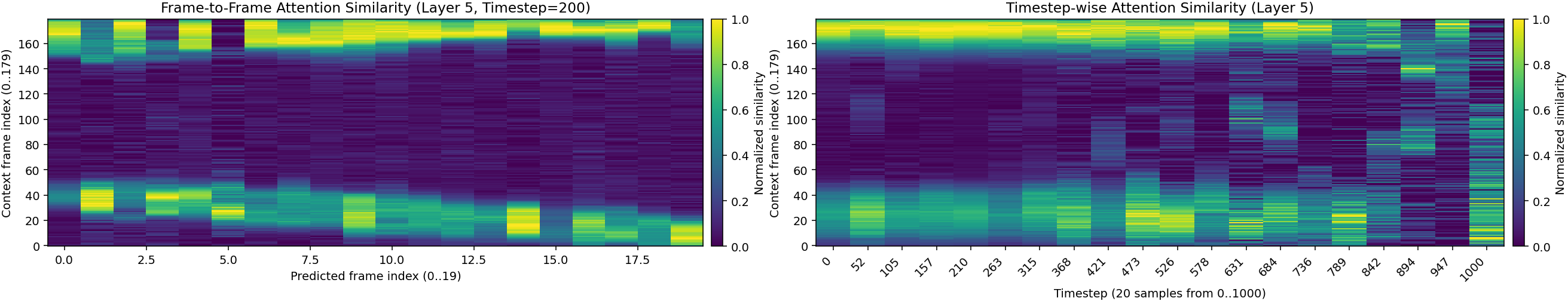}
  \hfill
  \includegraphics[width=0.49\linewidth]{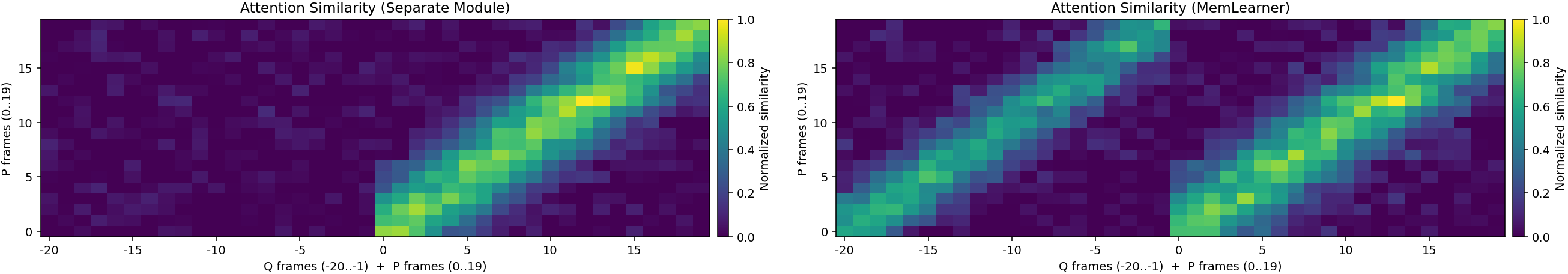}
  \caption{Attention visualization results.}
  \label{fig:attn_similarity}
\end{figure}

\paragraph{Adaptive Attention Across Frames and Timesteps.}
As shown in the left half of Fig.~\ref{fig:attn_similarity}, query tokens exhibit markedly different attention distributions across predicted frames and diffusion timesteps. Early diffusion timesteps attend more broadly to context tokens, capturing global scene layout, while later timesteps focus on fine-grained local correspondences. This confirms that different predicted frames and denoising stages require different information from context, justifying our adaptive query design over naive rule-based context retrieval.

\paragraph{Failure of the Separate Module.}
As shown in the right half of Fig.~\ref{fig:attn_similarity}, the separate context query module (Fig.~\ref{fig:arch_clarify}(b) in the main text) produces near-zero attention similarity between query and context tokens, indicating a failure to establish meaningful context modeling. This lack of effective information flow hinders gradient propagation to the video generation model, which consequently ignores the separate context query module's output and degrades to a text-to-video model, consistent with the quantitative results in Tab.~\ref{tab:comp} in the main text.

\subsection{Evaluation on Real-World SpatialVID Dataset}
\label{sec:spatialvid}

To evaluate generalization to real-world scenarios, we conduct quantitative experiments on the SpatialVID dataset~\cite{spatialvid}. We select SpatialVID for two reasons: (1) it contains long videos with sequences up to 900 frames, and (2) it provides per-frame camera pose annotations, enabling controlled evaluation. We randomly sample 1,000 long sequences from the dataset for testing, ensuring no overlap with training data.

As shown in Tab.~\ref{table:comp_spatialvid}, MemLearner consistently outperforms baselines across all metrics. However, the performance gaps are relatively small compared to those on our collected dataset (Tab.~\ref{tab:comp} in the main text). This is expected: SpatialVID filters out videos with large camera motions to ensure accurate pose estimation, and real-world videos inherently lack frequent revisit patterns (as illustrated in Tab.~\ref{tab:dataset_comparison} of the main text). These characteristics make SpatialVID a relatively easy benchmark for evaluating memory capability.

\begin{table}[h]
\centering
\begin{tabular}{l | c c | c c}
\toprule
& \multicolumn{2}{c|}{GT Comp.} & \multicolumn{2}{c}{Revisit Comp.}\\
Methods  & PSNR$\uparrow$ & LPIPS$\downarrow$ & PSNR$\uparrow$ & LPIPS$\downarrow$ \\
\hline\hline
VMem~\cite{vmem} & 21.76 & 0.2844 & 18.29 & 0.3301 \\
CaM~\cite{cam} & 21.83 & 0.2674 & 18.41 & 0.3285 \\
Ours & \textbf{22.46} & \textbf{0.2576} & \textbf{18.75} & \textbf{0.3078} \\
\bottomrule
\end{tabular}
\caption{Quantitative comparison on the real-world SpatialVID dataset~\cite{spatialvid}.}
\label{table:comp_spatialvid}
\end{table}

\subsection{User Study}
\label{sec:user_study}
In Table~\ref{table:user_study}, we also conduct a user study, and the results clearly show that users overwhelmingly prefer our method for both visual quality and scene consistency.
\input{subtex-fig-tex/tab_user_study}

\subsection{Additional Attention Computational Comparison}
\label{sec:attn_compute}
We provide a supplementary comparison of computational cost and performance when including context tokens as queries in 3D attention computations. As shown in Table~\ref{table:ablation_attention_supp} and Figure~\ref{fig:attention_ablation_supp}, incorporating context tokens as queries introduces substantial computational overhead while offering negligible or even degraded performance.
\input{subtex-fig-tex/fig_compute_settings}
\input{subtex-fig-tex/tab_attention_compute_comp_supp}

\subsection{Additional Dataset Ablation Results}
\label{sec:dataset_ablation}
Figure~\ref{fig:comp_dataset_supp} shows additional qualitative results on how incorporating real-world training data improves the realism of generated videos.

\subsection{Additional Comparison Results}
In Figure~\ref{fig:comparison_supp}, we provide additional qualitative comparisons across different methods.

\subsection{Comparison with a Separately Trained Query Module}
\label{sec:alternative_module}
A natural alternative to our end-to-end design is to train the context query module separately with explicit supervision. Following this direction (separate pre-training, auxiliary supervision, and progressive fine-tuning), we pre-train a separate query module with an L1 loss between the query (Q) and predicted (P) tokens, and then jointly fine-tune it together with the video generation model. As shown in Tab.~\ref{table:alternative}, this alternative remains far below MemLearner. We attribute this to the difficulty of designing explicit supervision for context querying: what constitutes ``correctly queried memory'' is unknown, so any hand-crafted supervision target is likely suboptimal. In contrast, MemLearner avoids this issue by learning what to query end-to-end from the diffusion loss alone.

\begin{table}[h]
\centering
\begin{tabular}{l | c c | c c | c}
\toprule
& \multicolumn{2}{c|}{GT Comp.} & \multicolumn{2}{c|}{Revisit Comp.} &\\
Setting  & PSNR$\uparrow$ & LPIPS$\downarrow$ & PSNR$\uparrow$ & LPIPS$\downarrow$ & fps$\uparrow$ \\
\hline\hline
Alternative & 16.56 & 0.4929 & 15.71 & 0.6057 & 0.48 \\
Ours & \textbf{21.23} & \textbf{0.2904} & \textbf{18.57} & \textbf{0.3230} & \textbf{0.54} \\
\bottomrule
\end{tabular}
\caption{Comparison with a separately trained query module using auxiliary L1 supervision and progressive fine-tuning.}
\label{table:alternative}
\end{table}

\subsection{Ablation on Query Token Initialization}
\label{sec:query_init}
We further investigate how the initialization of query tokens affects performance. By default, MemLearner initializes the Q tokens as randomly sampled noise, which aligns with the input distribution of diffusion models. We also try initializing the Q tokens from a copy of the noisy predicted (P) tokens. As shown in Tab.~\ref{table:query_init}, the two initialization strategies perform comparably. This is because the Q-to-P attention lets the query tokens learn what the predicted tokens need at each forward pass; the query guidance is thus acquired through attention rather than from the specific initial values, confirming that the initialization is not critical.

\begin{table}[h]
\centering
\begin{tabular}{l | c c | c c}
\toprule
& \multicolumn{2}{c|}{GT Comp.} & \multicolumn{2}{c}{Revisit Comp.}\\
Q Init.  & PSNR$\uparrow$ & LPIPS$\downarrow$ & PSNR$\uparrow$ & LPIPS$\downarrow$ \\
\hline\hline
Noisy P tokens & 21.17 & 0.2938 & 18.67 & 0.3227 \\
Noise (Ours) & 21.23 & 0.2904 & 18.57 & 0.3230 \\
\bottomrule
\end{tabular}
\caption{Ablation on query token initialization. Noise initialization and noisy-P-token initialization perform comparably.}
\label{table:query_init}
\end{table}

\subsection{Comparison with Geometry-Based Retrieval}
\label{sec:vrag}
To further compare against geometry/position-based retrieval, we evaluate VRAG~\cite{vrag}, which differs from FOV-based retrieval (e.g., CaM~\cite{cam}) by leveraging geometric position cues to alleviate the wall-occlusion problem. As shown in Tab.~\ref{table:vrag}, MemLearner still outperforms VRAG across all metrics, demonstrating the advantage of learning-based context query over geometry-based retrieval.

\begin{table}[h]
\centering
\begin{tabular}{l | c c | c c}
\toprule
& \multicolumn{2}{c|}{GT Comp.} & \multicolumn{2}{c}{Revisit Comp.}\\
Methods  & PSNR$\uparrow$ & LPIPS$\downarrow$ & PSNR$\uparrow$ & LPIPS$\downarrow$ \\
\hline\hline
VRAG~\cite{vrag} & 19.61 & 0.3782 & 17.22 & 0.4008 \\
Ours & \textbf{21.23} & \textbf{0.2904} & \textbf{18.57} & \textbf{0.3230} \\
\bottomrule
\end{tabular}
\caption{Comparison with the geometry-based retrieval method VRAG~\cite{vrag}.}
\label{table:vrag}
\end{table}

\subsection{Zero-Shot Transfer to Real-World Epic-Kitchens}
\label{sec:epickitchens}
To validate generalization to out-of-distribution real-world scenarios, we conduct a zero-shot transfer evaluation on the Epic-Kitchens dataset~\cite{epickitchens}, a large-scale egocentric video dataset that differs substantially from our training data. As shown in Tab.~\ref{table:epickitchens}, MemLearner consistently outperforms CaM~\cite{cam} and VMem~\cite{vmem}, confirming that our learned memory mechanism generalizes to real-world scenarios and is not overfitted to the specific occlusion and dynamic-object patterns of our rendered training data.

\begin{table}[h]
\centering
\begin{tabular}{l | c c | c c}
\toprule
Epic-Kitchens & \multicolumn{2}{c|}{GT Comp.} & \multicolumn{2}{c}{Revisit Comp.}\\
Methods  & PSNR$\uparrow$ & LPIPS$\downarrow$ & PSNR$\uparrow$ & LPIPS$\downarrow$ \\
\hline\hline
VMem~\cite{vmem} & 18.85 & 0.4208 & 16.94 & 0.4450 \\
CaM~\cite{cam} & 19.31 & 0.3940 & 17.28 & 0.4206 \\
Ours & \textbf{20.19} & \textbf{0.3114} & \textbf{18.35} & \textbf{0.3375} \\
\bottomrule
\end{tabular}
\caption{Zero-shot transfer comparison on the real-world Epic-Kitchens dataset~\cite{epickitchens}.}
\label{table:epickitchens}
\end{table}

\subsection{Additional Video Quality Evaluation: VBench and CLIP Similarity}
\label{sec:vbench}
Since PSNR/LPIPS primarily assess pixel-level consistency, we additionally evaluate video quality with VBench~\cite{vbench} and semantic consistency with CLIP similarity~\cite{clip}. We report five VBench dimensions, namely Background Consistency (BG), Temporal Flickering (TF), Motion Smoothness (MS), Aesthetic Quality (AQ), and Imaging Quality (IQ), together with CLIP similarity (CLIP). As shown in Tab.~\ref{table:vbench}, MemLearner consistently outperforms CaM~\cite{cam} and VMem~\cite{vmem} across all metrics, confirming that its advantage holds beyond pixel-level metrics.

\begin{table}[h]
\centering
\begin{tabular}{l | c c c c c | c}
\toprule
Method & BG$\uparrow$ & TF$\uparrow$ & MS$\uparrow$ & AQ$\uparrow$ & IQ$\uparrow$ & CLIP$\uparrow$ \\
\hline\hline
VMem~\cite{vmem} & 0.9618 & 0.9504 & 0.9638 & 0.5883 & 0.6357 & 0.3219 \\
CaM~\cite{cam} & 0.9630 & 0.9539 & 0.9647 & 0.5895 & 0.6442 & 0.3190 \\
Ours & \textbf{0.9684} & \textbf{0.9572} & \textbf{0.9661} & \textbf{0.5905} & \textbf{0.6491} & \textbf{0.3255} \\
\bottomrule
\end{tabular}
\caption{Additional video quality evaluation on VBench~\cite{vbench} dimensions (BG: Background Consistency; TF: Temporal Flickering; MS: Motion Smoothness; AQ: Aesthetic Quality; IQ: Imaging Quality) and CLIP similarity.}
\label{table:vbench}
\end{table}

%% file: subtex-fig-tex/tab_user_study.tex
\begin{table}[h]
\tabcolsep=0.15cm
\center

\begin{tabular}{
  >{\centering\arraybackslash}p{2.5cm} |
  *{5}{>{\centering\arraybackslash}p{1.8cm}}
}
\toprule
& DFoT~\cite{dfot} & FramePack~\cite{framepack} & VMem~\cite{vmem} & CaM~\cite{cam} & Ours \\
\hline
\hline
Quality (\%) & 20.79 & 14.53 & 49.00 & 54.13 & \textbf{69.51} \\
Consistency (\%) & 9.11 & 15.09 & 41.88 & 45.58 & \textbf{72.93}\\
\bottomrule
\end{tabular}
\caption{We randomly selected one video clip to be predicted from each of the 13 scenes and compared \textsc{MemLearner} with others. 27 users chose their preferred video from a randomly ordered set, with multiple selections allowed. The table shows user preference rates.}
\label{table:user_study}
\end{table}

%% file: subtex-fig-tex/fig_compute_settings.tex
 \begin{figure}[h]
  \centering
  \includegraphics[width=0.7\linewidth]{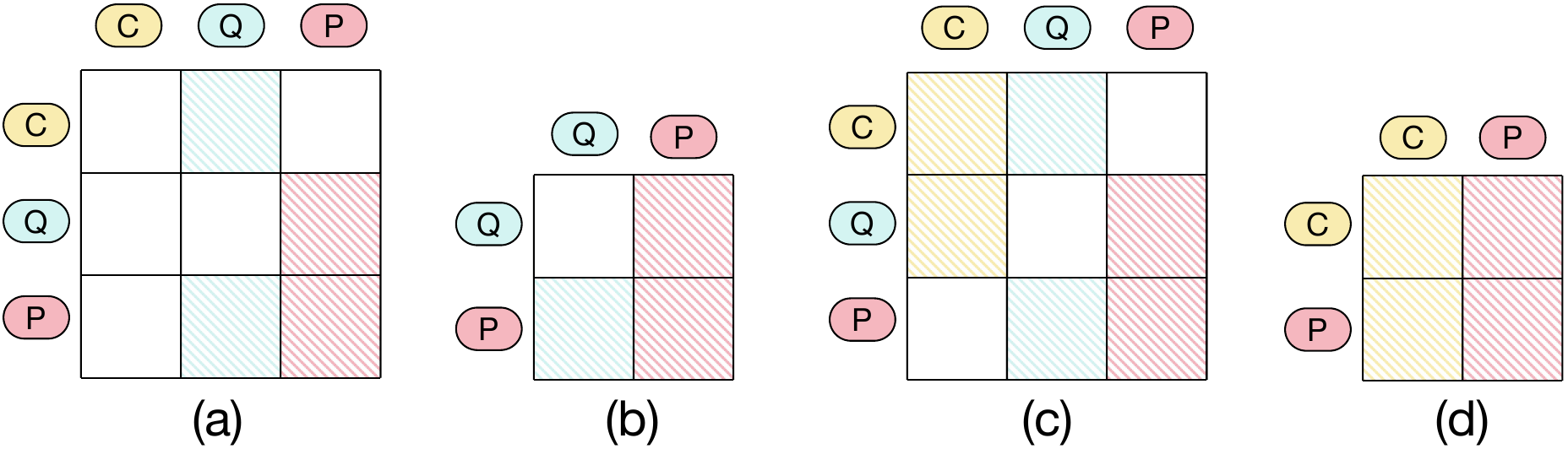}
  \caption{Additional attention computation settings. Horizontal tokens serve as queries; vertical tokens serve as keys/values. }
\label{fig:attention_ablation_supp} 
\end{figure}

%% file: subtex-fig-tex/tab_attention_compute_comp_supp.tex
\begin{table}[h]

\tabcolsep=0.09cm
\center

\begin{tabular}{c | c c | c c | c}
\toprule
& \multicolumn{2}{c}{GT Comp.} & \multicolumn{2}{c}{Revisit
 Comp.} &\\
  Setting  & PSNR$\uparrow$ & LPIPS$\downarrow$ & PSNR$\uparrow$ & LPIPS$\downarrow$  & Speed (fps)$\uparrow$ \\
  \hline
  \hline
   (a)*5+(b)*23 & 21.23 & 0.2904 & 18.57 & 0.3230 & \textbf{0.54} \\
(c)*5+(b)*23 & 21.15 & 0.2976 & 18.43 & 0.3291 & 0.46 \\
(c)*28 & 21.27 & 0.2914 & 18.62 & \textbf{0.3212} & 0.24 \\
(d)*28 & \textbf{21.34} & \textbf{0.2895} & \textbf{18.67} & 0.3227 & 0.28 \\

\bottomrule

\end{tabular}
\caption{Additional attention computation comparison. Settings (a)-(d) are shown in Fig.~\ref{fig:attention_ablation_supp}. The model has 28 layers total. In the notation, ``(a)*5+(b)*23'' indicates that the first 5 Query Layers use setting (a) and the remaining 23 Generative Layers use setting (b). This is the default setting for \textsc{MemLearner}. Other settings follow the same notation convention.}
\label{table:ablation_attention_supp}
\end{table}

%% file: tex/tab_fig.tex
\newpage

\input{subtex-fig-tex/fig_comp_dataset}
\input{subtex-fig-tex/fig_comp_supp}

%% file: subtex-fig-tex/fig_comp_dataset.tex
\begin{figure*}[t]
  \centering
  \includegraphics[width=1\linewidth]{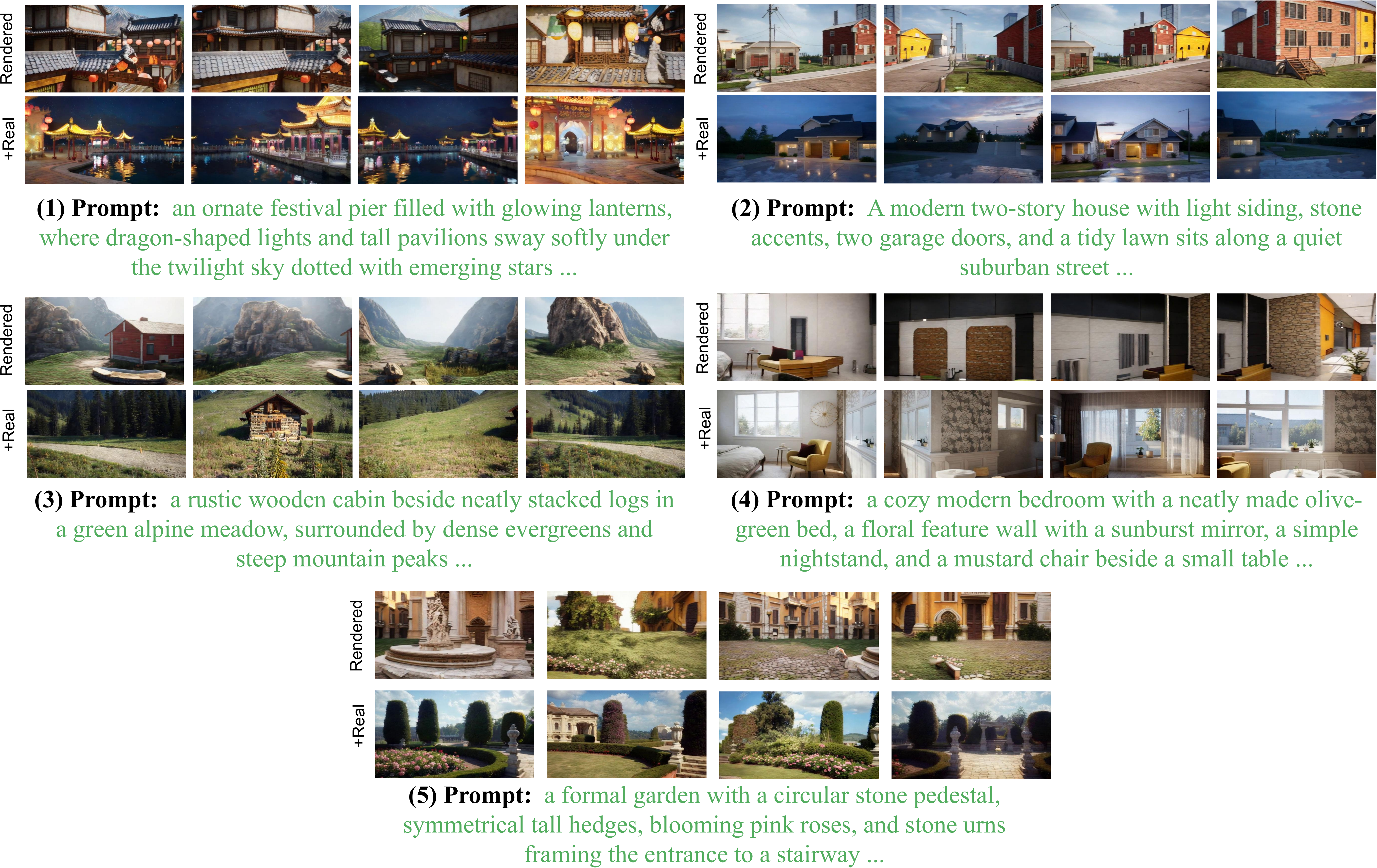}
  \caption{Additional Dataset Ablation Results. ``Rendered'' denotes results trained only on rendered videos, while ``+Real'' indicates results after adding real-world videos to the training set.}

\label{fig:comp_dataset_supp} 
\end{figure*}

%% file: subtex-fig-tex/fig_comp_supp.tex
\begin{figure*}[t]
  \centering
  \includegraphics[width=1\linewidth]{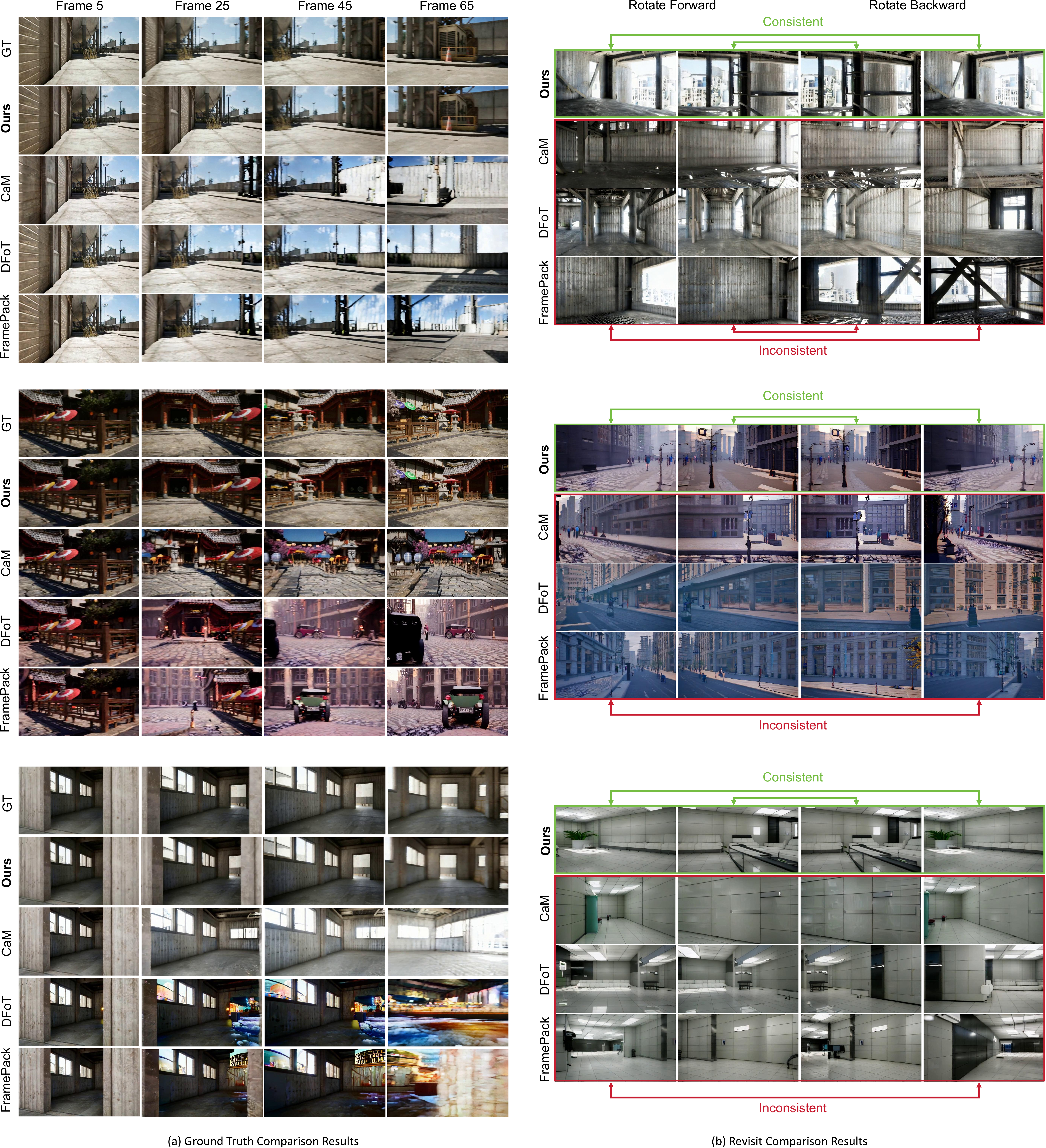}
  \caption{Additional Qualitative Comparison Results.}
\label{fig:comparison_supp} 
\end{figure*}